\documentclass[journal]{IEEEtran}
\ifCLASSINFOpdf
\else
\fi
\usepackage{microtype}
\usepackage{graphicx}
\usepackage{subfigure}
\usepackage{booktabs}
\usepackage{hyperref}
\usepackage{amssymb}
\usepackage{amsfonts}
\usepackage{MnSymbol}
\usepackage{multirow}
\usepackage{amsmath}
\usepackage{bm}
\usepackage{color}
\usepackage{array}
\newcolumntype{C}[1]{>{\centering}p{#1}}
\newtheorem{theorem}{Theorem}
\newtheorem{lemma}{Lemma}

\newtheorem{remark}{Remark}
\newtheorem{corollary}{Corollary}
\begin{document}
\title{Maximizing Conditional Independence for Unsupervised Domain Adaptation}
\author{Yi-Ming Zhai,
        You-Wei Luo
\thanks{Y.M. Zhai and Y.W. Luo are with the Intelligent Data Center, School
of Mathematics, Sun Yat-Sen University, Guangzhou, 510275, China.}
}
\markboth{Journal of \LaTeX\ Class Files,~Vol.~14, No.~8, August~2015}
{Shell \MakeLowercase{\textit{et al.}}: Bare Demo of IEEEtran.cls for IEEE Journals}
\maketitle
\begin{abstract}
Unsupervised domain adaptation studies how to transfer a learner from a labeled source domain to an unlabeled target domain with different distributions.
Existing methods mainly focus on matching the marginal distributions of the source and target domains, which probably lead a misalignment of samples from the same class but different domains.
In this paper, we deal with this misalignment by achieving the class-conditioned transferring from a new perspective. We aim to maximize the conditional independence of feature and domain given class in the
reproducing kernel Hilbert space. The optimization of the conditional independence measure can be viewed as minimizing a surrogate of a certain mutual information between feature and domain.
An interpretable empirical estimation of the conditional dependence is deduced and connected with the unconditional case.
Besides, we provide an upper bound on the target error by taking the class-conditional distribution into account, which provides a new theoretical insight for most class-conditioned transferring methods.
In addition to unsupervised domain adaptation, we extend our method to the multi-source scenario in a natural and elegant way.
Extensive experiments on four benchmarks validate the effectiveness of the proposed models in both unsupervised domain adaptation and multiple source domain adaptation.
\end{abstract}

\begin{IEEEkeywords}
Conditional independence, kernel method, domain adaptation, class-conditioned transferring.
\end{IEEEkeywords}

\IEEEpeerreviewmaketitle
\section{Introduction}
\IEEEPARstart{A}{lgorithms} of supervised learning have made tremendous contributions to artificial intelligence and have wide applications in real-life. Sufficient labeled data play a significant role in supervised learning. However, it is often expensive and time-consuming to collect plenty of labeled data. In contrast, it is much easier to collect considerable unlabeled data. An intuitive idea is to apply the learned predictive model, which has been trained with the labeled data in a supervised way, to the unlabeled dataset directly. However, there may exist a large discrepancy between the training and testing sets due to the existence of dataset shift \cite{long2018transferable}. An direct application may result in a degradation of recognition performance.

Unsupervised Domain Adaptation (UDA) is proposed to deal with this degradation by learning a discriminative predictor in the presence of a labeled source domain and an unlabeled target domain \cite{pan2009survey}. The source and target domains have similar but not identical distributions due to different domain-specific information, such as image styles, camera views, illuminations and backgrounds \cite{pan2010domain,Ren_TSCDA}. Exploring the invariant representations across domains is vital for UDA. Recently, UDA is receiving more attention due to its widespread applications in object recognition \cite{lin2016cross},
object detection \cite{Khodabandeh_2019_ICCV}, speech recognition \cite{Khur2021}, disease diagnosis~\cite{Xu_FLARE} and so on, which promote the growth of industry greatly.

\begin{figure}[t]
\vskip 0.15in
\begin{center}
\includegraphics[width=238pt, trim= 0 30 0 50]{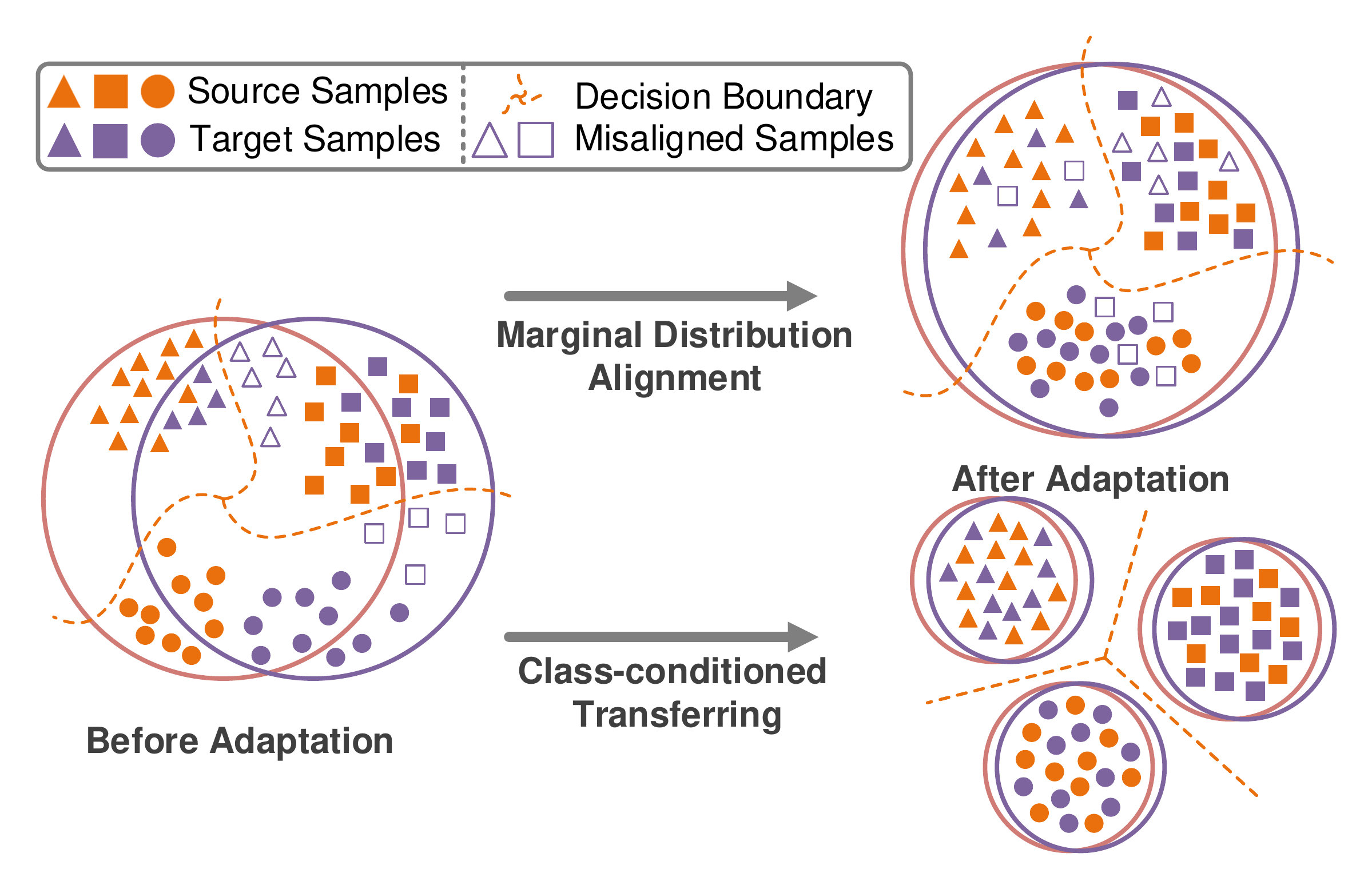}
\caption{Illustration of the class-conditioned transferring. Most methods focus on the
marginal distribution alignment, which may lead a misalignment. The class-conditioned
transferring deals with this misalignment by exploring the class-conditional distributions across domains. Best viewed in color.}
\label{fig: UDA}
\end{center}
\vspace{-0.26in}
\end{figure}

Covariate shift assumes the source and target domains have different feature distributions though share the same feature space, \textit{i.e.}, $P_X^s \neq P_X^t$ with $P_{Y|X}^s = P_{Y|X}^t$. Inspired by the rigorous transfer theory \cite{ben2007analysis}, various UDA methods have been proposed to learn domain-invariant representations by minimizing the discrepancy of the marginal distributions. Statistical approaches attempt to minimize the distribution
discrepancy by aligning the moment statistics in the kernel embedding space \cite{long2013transfer}.
In literature \cite{pan2010domain}, the authors explore a latent feature space
where the domain distribution discrepancy is minimized by Maximum Mean
Discrepancy (MMD).
Manifold learning frameworks \cite{gong2012geodesic,ren2019heterogeneous,luo2020unsupervised} consider the geodesic distance between domains after mapping the raw data into Riemannian manifolds or subspaces. Recently, deep learning methods have achieved remarkable performance in alleviating dataset shift due to the powerful nonlinear transformation, generalization ability and fitting ability \cite{yosinski2014transferable}. The success of adversarial adaptation methods \cite{ganin2016domain,tzeng2017adversarial,ren2019domain} have shown huge potential in generating domain-indistinguishable representations and forming a well-aligned marginal distribution.

Though previous works have achieved considerable progresses in UDA, there still exist bottlenecks as these methods mainly focus on matching marginal distributions and ignore the discriminative structures among samples from different classes. Samples from the same class but different domains may not be mapped nearby in the latent feature space, even with a perfect alignment of marginal distribution, which is described in Figure \ref{fig: UDA}(up).
With the motivation of learning a domain-invariant and discriminative classifier, class-conditioned transferring has been proposed and explored recently~\cite{pan2019transferrable,liang2019distant,Luo_2021_CVPR}.
As shown in Figure \ref{fig: UDA}(down), a more accurate class-conditioned alignment will promote a more accurate domain alignment, which is encouraging for following tasks.

Multi-source domain  adaptation (MDA) aims to transfer knowledge from multiple source domains to the unlabeled target domain. MDA is more practical as training samples may be gathered from multiple sources with different distributions \cite{sun2015survey}. However, the above UDA methods are specifically proposed for the single-source scenario. Simply combining different source domains into one source domain and directly applying the above UDA methods may lead a suboptimal solution, since the discrepancy among source domains is ignored and the data from different source domains may interfere with each other during the learning process \cite{riemer2018learning}. Therefore, effective MDA algorithms are required to deal with the increased dataset shift.

To address the aforementioned bottlenecks, in this paper, we propose a novel method called Maximizing Conditional Independence (MCI) for unsupervised domain adaptation. The key difference over previous UDA methods is that MCI aims to remove the domain-specific information by achieving the conditional independence. It is a totally new statistical perspective to deal with dataset shift in domain adaptation. More precisely, MCI models the sets of variables, \textit{i.e.}, extracted feature $X$, class $Y$, and domain $Z$ at the same time, and then exploits the normalized conditional cross-covariance operator in Reproducing Kernel Hilbert Space (RKHS) to remove the domain-specific information from the class-conditioned feature space. From the perspective of information theory, MCI seeks a compact and informative feature space with reduced class-conditioned mutual information between feature and domain. As domain is also modeled along with the extracted feature, MCI can be extended to deal with MDA problem naturally, which leads wider applications in practical scenarios.

To the best of our knowledge, maximizing the conditional independence for class-conditioned transferring has not been explored in domain adaptation.
The contributions of our work are mainly summarized as:
\begin{itemize}
\item[1)] We provide class-conditional distribution based generalization error bound under UDA and MDA scenarios, which gives a new theoretical insight for most class-conditioned transferring methods.
\item[2)] We propose a simple yet effective method MCI for UDA, which achieves the class-conditioned transferring by making feature and domain conditionally independent given class. It can also be viewed as minimizing a surrogate of a certain mutual information. Additionally, MCI is extended as MS-MCI for MDA.
\item[3)] We mathematically derive that the conditional independence will lead a class-conditional distribution alignment, which guarantees that samples from the same class but different domains are mapped nearby in the latent feature space.
\item[4)] We derive an interpretable empirical estimation of the conditional dependence and connect it with the corresponding estimation in the unconditional case, which adjusts and improves the results in \cite{fukumizu2007kernel}.
\end{itemize}

The rest of this paper is organized as follows. Related works about UDA are reviewed briefly in Section~\ref{Related Work}. Section~\ref{method} provides some preliminaries about the conditional dependence measure, details of MCI, extension for MDA and theoretical analysis. Extensive experiments along with analysis under UDA and MDA scenarios are presented in Section~\ref{Experiments}. Finally, Section~\ref{conclusion} concludes this paper.

\section{Related Work}
\label{Related Work}
Domain adaptation has gained more and more attention due to its wide applications, and has been widely explored in the past years. In this section, we give a brief desciption about UDA and MDA.

\textit{1) Unsupervised Domain Adaptation:}
Most methods of domain adaptation aim to reduce the discrepancy across domains and learn domain-invariant features under the covariate shift assumption.
Moment matching based method CORrelation ALignment (CORAL) \cite{sun2016return} minimizes the dataset shift by aligning the second-order statistics of the source and target distributions.
Compared with MMD, weighted MMD (WMMD) \cite{Yan_2017_CVPR} takes class prior distribution of the source domain into account, which provides a better metric for domain discrepancy.
Geodesic Flow Kernel (GFK) \cite{gong2012geodesic} and Discriminative Manifold Propagation
(DMP) \cite{luo2020unsupervised} tackle the covariate shift from the manifold alignment perspective.
Courty  \textit{et al.} \cite{courty2016optimal} aim to address the dataset shift by
learning the nonlinear Wasserstein map across domains from an optimal transport perspective.
Enhanced Transport Distance (ETD) \cite{li2020enhanced} builds an attention-aware transport distance to
measure the domain discrepancy. Kernel Gaussian-Optimal Transport Map (KGOT) \cite{zhang2019optimal}
matches the distributions by formulating the optimal transport problem in RKHS
with Gaussian prior.
Adversarial adaptation method Domain-adversarial Neural Network (DANN) \cite{ganin2016domain}
aims to learn domain-invariant representations by training a domain discriminator and a domain-indistinguishable feature extractor.
Deep Adaptation Networks (DAN) \cite{long2018transferable} matches the marginal distributions by embedding the deep features into the RKHSs.

Though these methods have aligned the marginal distributions, they ignore the negative transferring among different classes, which may reduce the classification accuracy in
real-life scenarios.
Thus, performing the class-conditioned adaptation is necessary to improve the discriminability of the features.
Conditional Domain Adversarial Networks (CDANs) \cite{long2018conditional}
condition the discriminative information conveyed in the classifier predictions, which aims to match the joint distribution of feature and class.
Jiang \textit{et al.} \cite{jiang2020implicit} propose an implicit class-conditioned domain alignment framework from a sampling perspective.
Several works have been proposed to  minimize the divergence of class-conditional clusters or prototypes across domains.
Moving Semantic Transfer Network (MSTN) \cite{xie2018learning} for UDA learns semantic representations for unlabeled target samples by aligning the labeled source centroid and the pseudo-labeled target centroid.
Cluster Alignment with a Teacher (CAT) \cite{deng2019cluster} exploits the discriminative class-conditional structures of distributions by aligning the corresponding clusters across domains.
However, these methods cannot promise the class-conditional distribution alignment theoretically.
Zhao \textit{et al.} \cite{zhao2019learning} propose an information-theoretic generalization bound which shows that matching the class-conditional distributions is nonnegligible.
Deep Subdomain Adaptation Network (DSAN) \cite{zhu2020deep} presents a deep transfer network to match the conditional distributions based
on the local maximum mean discrepancy (LMMD).
A kernel learning method (KLN) \cite{Ren2021} simultanously learn a more expressive kernel and label prediction distribution, which enhances the effectiveness of CMMD. Conditional Kernel Bures \cite{Luo_2021_CVPR} aims to seek a kernel covariance embedding for characterizing conditional distribution discrepancy. Differently, MCI aims to remove domain-specific information by maximizing conditional independence innovatively, which simultaneously bring a class-conditional distribution alignment.

Previous work MIDA \cite{Ke2018} investigates the independence between feature and domain, which aims to match the marginal distributions of domains.
Differently, MCI seeks the conditional independence between feature and domain given class, which leads a class-conditional distribution alignment of $P_{X|Y}^s$ and $P_{X|Y}^t$. Moreover, MCI can be viewed as the reduction of the shared information between feature $X$ and domain $Z$ conditionally given class $Y$.

\textit{2) Multi-source Domain Adapation:} MDA aims to transfer knowledge from multi-source domains to an unlabeled target domain. Most UDA algorithms only focus on the single-source scenario, though massive labeled source samples may be gathered from multiple domains with different distributions in many practical scenarios.
Thus, it's necessary and valuable to find an efficient algorithm for MDA.

There are some theoretical analyses for MDA based on \cite{blitzer2007learning,ben2010theory}. Mansour \textit{et al.} \cite{mansour2009domain} assume that target distribution can be represented by a combination of weighted source domain distributions. Thus, a perfect target classifier can be obtained by weighted combination of the source classifiers. Motivated by this combination rule, Xu \textit{et al.}~\cite{xu2018deep} learn domain-invariant features by proposing Deep Cocktail Network (DCTN) based on multi-way adverarial learning. Liu \textit{et al.}~\cite{Liu_TWMDA} align the target and multiple sources at domain-level by an adversarial learning process, and further reduce the domain gap at class-level by minimizing the distance between the class prototypes and unlabeled target instances. Ren \textit{et al.}~\cite{PTMDA_Ren2022} explore the structured domain-invariant information by iteratively mapping each group of source and target domains into a group-specific subspace, which is learned in an adversarial manner. Multiple Domain Matching Network (MDMN) \cite{li2018extracting} embeds all samples into a shared feature space while learning which domains share strong statistical relationships based on the Wasserstein-like measure.
Peng \textit{et al.} \cite{peng2019moment} provide a theoretical analysis for the moment matching approaches, and propose Multi-Source Domain Adaptation ($\text{M}^3$SDA) that dynamically aligns multiple domains by matching the moments of the features distributions.
Zhao \textit{et al.} \cite{zhao2018adversarial} give task-adaptive generalization bounds and propose Multi-source Domain Adversarial Networks (MDAN) based on the theoretical results. Wen \textit{et al.} \cite{wen2020domain} provide a finite-sample generalization bound based on the the domain discrepancy, and accordingly propose Domain AggRegation Network (DARN) dynamically to adjust the importance weights of he source domains during the course of training. Zhu \textit{et al.}~\cite{zhu2019aligning} propose Multiple Feature Spaces Adaptation Network (MFSAN), which simultaneously learns domain-invariant representations and aligns the classifiers' outputs for target samples.

Compared with these elaborate MDA methods, our MS-MCI captures the discriminative structure behind different conditional distributions from a conditional independence view, which also leads a class-conditional distribution alignment across domains. Additionally, it is rather direct and intuitive to generalize MCI from UDA to MDA.

\section{Methodology}
\label{method}
In this section, we firstly introduce conditional dependence measure, and relate it with conditional independence. Secondly, we present that employing the conditional independence aims to remove the domain-specific information. Then, we propose an interpretable empirical estimation of conditional dependence. Next, we discuss the objective of MCI and provide theoretical analysis. Finally, we extend MCI as MS-MCI for multi-source scenario and give a further discussion from a theoretical insight.
\subsection{Measuring Conditional Independence in RKHS}
Given random variables $X$, $Z$, $Y$ on $\mathcal{X}$, $\mathcal{Z}$, $ \mathcal{Y}$, respectively. The RKHSs of functions on $\mathcal{X}$, $\mathcal{Z}$ and $\mathcal{Y}$ are denoted by $\mathcal{F}_{\mathcal{X}}$, $\mathcal{F}_{\mathcal{Z}}$ and $\mathcal{F}_{\mathcal{Y}}$, respectively. $X \upmodels Z \mid Y$ means that $X$ and $Z$ are conditionally independent given $Y = y$, $\forall y \in \mathcal{Y}$. $\mathcal{N}\left(T\right)$ and $\mathcal{R}\left(T\right)$ refer to the null space and the range of an operator $T$, respectively.

The \textit{cross-covariance operator} $\Sigma_{ZX}$ from $\mathcal{F_{\mathcal{X}}}$ to $\mathcal{F_{\mathcal{Z}}}$
is defined to satisfy
\begin{equation*}
\left\langle g, \Sigma_{Z X} f\right\rangle_{\mathcal{F}_{\mathcal{Z}}}= \mathbb{E}[f(X) g(Z)]-\mathbb{E}[f(X)] \mathbb{E}[g(Z)]
\end{equation*}
for all $f \in \mathcal{F}_{\mathcal{X}}$ and $g \in \mathcal{F}_{\mathcal{Z}}$.

It is known that $\Sigma_{Z X}$ can be represented by the covariance of the marginals and the correlation
\cite{baker1973joint}. Thus, we have
\begin{equation}\label{SigamaZX}
\Sigma_{ZX} = \Sigma_{ZZ}^{\frac{1}{2}} V_{ZX} \Sigma_{XX}^{\frac{1}{2}},
\end{equation}
where $V_{ZX}$ is a unique bounded operator, called the \textit{normalized cross-covariance operator}
(NOCCO).
Since $V_{ZX}$ is normalized, it satisfies that
$\mathcal{R}\left(V_{ZX}\right)\subset \overline{\mathcal{R}\left(\Sigma_{ZX}\right)} $,
and $\mathcal{N}\left(V_{ZX}\right)^{\perp} \subset \overline{\mathcal{R}\left(\Sigma_{ZX}\right)}$ \cite{fukumizu2007statistical}.

The \textit{normalized conditional cross-covariance operator} (COND) \cite{fukumizu2007kernel} is defined as
\begin{equation*}
V_{ZX \mid Y} = V_{ZX} - V_{ZY}V_{YX},
\end{equation*}
where $V_{ZY}$ and $V_{YX}$ can be derived by Eq.~\eqref{SigamaZX}. This operator is defined for measuring the conditional dependence of random variables $X$ and $Z$
given $Y$.

We denote $\widetilde{X} = \left(X,Y\right)$, $\widetilde{Z} = \left(Z,Y\right)$ with the kernel product $k_{\widetilde{\mathcal{X}}} = k_{\mathcal{X}} k_{\mathcal{Y}}$ and $k_{\widetilde{\mathcal{Z}}} = k_{\mathcal{Z}} k_{\mathcal{Y}}$.
Corollary 9 in \cite{fukumizu2004dimensionality} implies that $X \upmodels Z \mid Y$
if and only if $\Sigma_{\widetilde X \widetilde Z \mid Y} = \textit{0}$.
With the distribution embedding property of RKHS, the following lemma
connects independence to operators $V_{X Z}$ and $ V_{\widetilde X \widetilde Z \mid Y}$.

\begin{lemma}[\cite{fukumizu2007kernel}, Theorem 3]
\label{Lemma: independence with operators}
(i) If the product $k_{\mathcal{X}} k_{\mathcal{Z}}$ is characteristic, then
\begin{align*}
X \upmodels Z &  \Longleftrightarrow V_{X Z} = \textit{0}.
\end{align*}
(ii) Assume that the product $k_{\widetilde{\mathcal{X}}} k_{\widetilde{\mathcal{Z}}}$ is a characteristic kernel on $\mathcal{X} \times \mathcal{Y}\times \mathcal{Z}$,
and $\mathcal{F}_{\mathcal Y} + \mathbb{R}$ is dense in $L^2(P_{\mathcal{Y}})$. Then,
\begin{align*}
X \upmodels Z \mid Y  \Longleftrightarrow V_{\widetilde{X} \widetilde{Z}| Y} = \textit{0}.
\end{align*}
\end{lemma}

Note $\mathcal{F}_{\mathcal Y} + \mathbb{R}$ is dense in $L^2(P_{\mathcal{Y}})$ means that
$k_{\mathcal{y}}$ is bounded and characteristic \cite{fukumizu2009kernel}.
To measure the distance between the zero element \textit{0} and $V_{XZ}$ ($V_{\widetilde{X} \widetilde{Z}| Y}$), the HS norm $\| \cdot \|_{HS}$ of operators is employed.
Denote that $V: \mathcal{F}_1 \rightarrow \mathcal{F}_2$ is a linear operator, $\{\phi_{i}\}$ and $\{ \psi_{j} \}$ are complete orthonormal systems (CONSs) of
$\mathcal{F}_1$ and $\mathcal{F}_2$ \cite{gretton2005measuring}.
The HS norm of $V$  is defined as
$\| V\|_{HS}^2=\Sigma_{i,j} \langle \psi_{j}, V\phi_{i} \rangle_{\mathcal{F}_2}^{2}$.
$V$ is a HS operator if
the sum $\Sigma_{i,j} \langle \psi_{j}, V\phi_{i} \rangle_{\mathcal{F}_2}^{2}$ is finite.
As shown in \cite{fukumizu2007kernel}, $V_{X Z}$ and $V_{\widetilde Y \widetilde X|Z}$ are HS operators, thus we can measure the statistical dependence as:
\begin{align*}
I^{NOCCO}\left( X, Z \right)& = \| V_{ Z X } \|_{HS}^{2},\\
I^{COND}\left( X, Z|Y \right)& = \| V_{\widetilde Z \widetilde X \mid Y} \|_{HS}^{2}.
\end{align*}

\subsection{Removing Domain-Specific Information}
\label{Domain Private Information Removing}

To remove domain-specific information and learn class-conditioned domain-invariant representations, we consider the multiplicative interactions among feature, domain and class in RKHSs by employing COND.

We consider a feature space $\mathcal{X}$, a label space $\mathcal{Y}$, and a domain label space $\mathcal{Z}$ for the source and target domains.
Denote the random variables $X \in \mathbb{R}^d$, $Y \in \{ \mathbf{y}_1,\ldots, \mathbf{y}_K \}$, $Z \in \{\mathbf{z}^s, \mathbf{z}^t\}$, where $\mathbf{z}^s$ represents the source domain, $\mathbf{z}^t$
the target domain, $\mathbf{y}_k$ the class $k$.
Samples from class $\mathbf{y}_k$ but different domains can be represented as $X^s_k \sim P(X|Z = \mathbf{z}^s,Y = \mathbf{y}_k)$ and
$X^t_k \sim P(X|Z = \mathbf{z}^t,Y = \mathbf{y}_k)$.

\begin{figure}[t]
\vskip 0.2in
\begin{center}
\centering{\includegraphics[width=180pt, trim= 300 180 600 270]{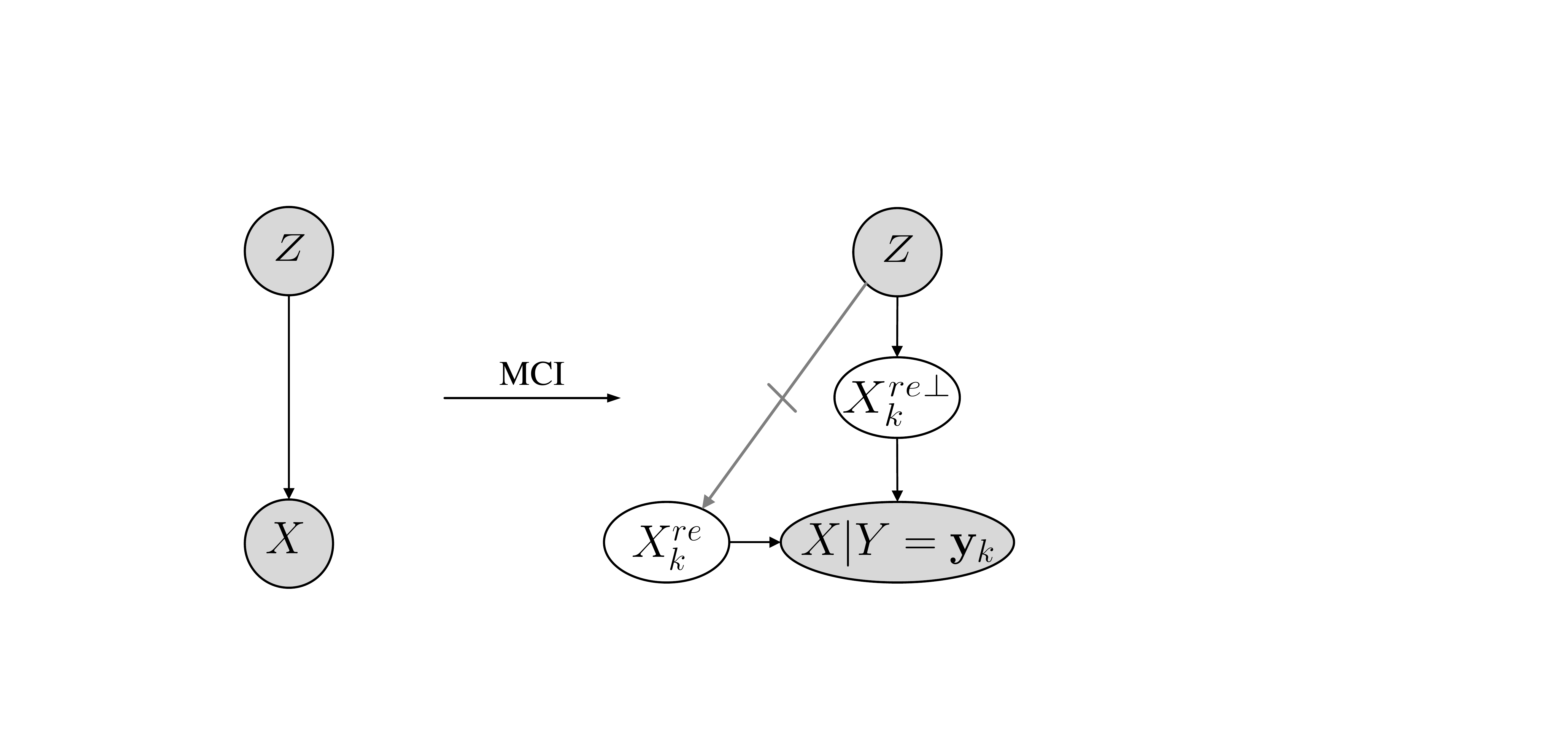}}
\caption{A directed graph of MCI. $X^{re}_k$ denotes the class-conditioned domain-invariant features, which is the output of the feature transformation $g(\cdot)$. ${X^{re}_k}^{\perp}$
denotes the remaining features depending on the domain label $Z$. Some methods aims to make $X$ independent with $Z$. Differently, MCI aims to find $X^{re}_k$ being conditionally independent of $Z$ given class $Y$.}
\label{fig: COND_loss}
\end{center}
\vskip -0.2in
\end{figure}

As shown in the left of Figure \ref{fig: COND_loss}, feature $X$ is conditioned on domain $Z$. Thus, source and target domains are supposed to have different marginal distributions.
Most UDA methods tend to mitigate the influence of domain-specific information by minimizing domain discrepancy, which may lead a misalignment across classes.
Differently, MCI explores the class-conditioned transferring by considering $Z \rightarrow X|Y=\mathbf{y}_k$, which is shown in the right of Figure~\ref{fig: COND_loss}.
It is obvious that feature $X$ and domain $Z$ are not conditionally independent,
\textit{i.e.},
\begin{equation*}
P(X,Z|Y) \neq P(X|Y)P(Z|Y),
\end{equation*}
which can be better understood by
\begin{equation}
\label{Eq: conditional independence}
P(X|Z,Y) \neq P(X|Y).
\end{equation}
Thus, $X^s_k$ and $X^t_k$ have similar but not identical conditional distributions
due to the domain-specific information.
In order to remove the domain-specific information from the class-conditioned feature space, we decompose the class-conditioned feature space with the direct product $X_k=X_k^{re} \oplus {X_k^{re}}^{\perp}$,
where ${X_k^{re}}^{\perp}$ is the orthogonal complement of $X^{re}_k$ and ${X_k^{re}}^{\perp}$  contains all the domain-specific information related to $Z$.
Thus, $X^{re}_k$ is independent of domain $Z$.
Conditioning the whole sample space of $Y$, we will have the conditional independence of feature $X$ and domain $Z$ given class $Y$, \textit{i.e.}, $X^{re} \upmodels Z \mid Y$.

The key to remove the domain-specific information and obtain class-conditioned domain-invariant representations is achieving the conditional independence of $X^{re}$ and $Z$ given $Y$.
We exploit the COND w.r.t. $X^{re} = g(X)$ in RKHSs to reduce the conditional dependence, where $g(\cdot)$ is a feature transformation.
Denote the extended variables as $\widetilde X = (X^{re}, Y)$ and $\widetilde Z = (Z, Y)$.
According to Lemma~\ref{Lemma: independence with operators}, minimizing $\| V_{\widetilde Z \widetilde X \mid Y} \|_{HS}^{2}$
is equal to learning a feature transformation $g(\cdot)$ which makes $X^{re}$
and $Z$ conditionally independent given $Y$,
\begin{equation}
\label{Eq: map h}
\min\limits_g \| V_{\widetilde Z \widetilde X \mid Y} \|_{HS}^{2}.
\end{equation}

We also provide an insight to MCI from an information theory perspective.
Mutual information is used to measure the information shared between two random variables.
More precisely, mutual information $I(X_k,Z) = H(X_k)-H(X_k|Z)\geqslant 0$ with equality if, and only if, $X_k$ and $Z$ are independent.
We have the inequality $I(X_k,Z) \leqslant I^{NOCCO}(X_k,Z)$ holds under the assumption of Theorem 4 in \cite{fukumizu2007kernel}.
In some way, optimizing MCI is equal to applying NOCCO on each class separatively.
Therefore, our MCI can be viewed as a surrogate for minimizing a certain mutual information $I(X_k^{re},Z)$ for each class theoretically.
From the perspective of mutual information, MCI is equal to reducing the shared information between feature $X$ and domain $Z$ conditionally given class $Y$.
Unfortunately, mutual information cannot be adaopted to remove the domain-specific information, since the direct estimation of mutual information is intractable if the joint distribution is highly complex.
Differently, the empirical estimation of the conditional dependence can be measured in the kernel space directly, without estimating any distributions.

Following theorem relates the conditional independence to class-conditional distribution alignment.
\begin{theorem}
\label{Theorem: conditional distribution alignment}
Assume that the product $k_{\mathcal{\widetilde{X}}}k_{\mathcal{Z}}$ is a characteristic kernel on $\mathcal{X}\times\mathcal{Y}\times\mathcal{Z}$, and $\mathcal{F}_{\mathcal{Y}} + \mathbb{R}$ is dense in $L^2(P_{\mathcal{Y}})$. For any conditional distributions $P_{X|Y}^{S}$, $P_{X|Y}^{T} \in {\rm Pr}^{S}(\mathcal{X}|\mathcal{Y})$, we have
\begin{equation*}
V_{\widetilde X \widetilde Z |Y} = 0 \quad \Longrightarrow \quad P_{X^{re}|Y}^S = P_{X^{re}|Y}^T.
\end{equation*}
\end{theorem}

According to Theorem~\ref{Theorem: conditional distribution alignment}, if the conditional dependence objective Eq.~\eqref{Eq: map h} is zero, the feature $X^{re}$ is class-conditioned
domain-invariant, \textit{i.e.}, $P_{X^{re}|Y}^S = P_{X^{re}|Y}^T$. More precisely, the conditional distribution of $X^{re}$ is essentially and solely determined by class $Y$, and domain $Z$ will be
superfluous once $Y$ is given, \textit{i.e.},
\begin{equation*}
P(X^{re}|Z = \mathbf{z}^s,Y = \mathbf{y}_k) = P(X^{re}|Z = \mathbf{z}^t,Y = \mathbf{y}_k).
\end{equation*}

Here we obtain a desired conclusion that our COND based method not only achieves the class-conditioned transferring but also derives a class-conditional distribution alignment.
By removing the domain-specific information while preserving the identical discriminative information, samples from the same class but different domains are supposed to be indistinguishable in the class-conditioned feature space.

\subsection{Empirical Estimation of the HS Measures}
\label{Empirical Estimates of the HS Measures}
In order to learn the feature transformation $g(\cdot)$ with finite data, we provide an empirical estimation of $\| V_{\widetilde Z \widetilde X \mid Y} \|_{HS}^{2}$ in the following.

Given a source domain $\mathcal{D}^S=\{( \bm{x}_i^s , \bm{y}_i^s )\}_{i=1}^{n_s}$, and a target domain $\mathcal{D}^T=\{\bm{x}_j^t \}_{j=1}^{n_t}$, where $\bm{x}_i^s$, $\bm{x}_j^t \in \mathcal{X}$ and ${\bm{y}_i^s \in \mathcal{Y}}$. Let $\mathbf X \in \mathbb{R}^{d \times n}$ be the $d$-dimensional feature matrix of $n$ samples from the source and the target domains, where $n = n_s+n_t$. With the feature transformation $g(\cdot)$, we have $\mathbf{X}^{re} \in \mathbb{R}^{d' \times n}$. $\mathbf Y \in \mathbb{R}^{K \times n}$ and $\mathbf Z \in \mathbb{R}^{2 \times n}$ are corresponding class label matrix and domain label matrix.
As the probability predictions of the classifier $C(\cdot)$ pre-trained on the source domain are mostly correct, it is natural to label the target samples by $\hat {\mathbf{Y}}^t$. The kernel feature maps $\phi(\cdot)$, $\lambda(\cdot)$ and $\psi(\cdot)$ are used to map $\widetilde{\mathbf X }$, $\widetilde{\mathbf Z }$ and $\mathbf{Y}$ into RKHSs, where $\widetilde{\mathbf X } = (\mathbf {X}^{re};\mathbf{Y}) \in \mathbb{R}^{(d'+K) \times n} $, $\widetilde{\mathbf Z} = (\mathbf {Z};\mathbf{Y}) \in \mathbb{R}^{(2+K) \times n} $. Then, we obtain kernel matrices
\begin{equation*}
\begin{aligned}
\mathbf{K}_{\widetilde{\mathbf X}} = {\mathbf\Phi}^T {\mathbf\Phi},
~~~~\mathbf{K}_{\widetilde{\mathbf Z}} = {\mathbf\Lambda}^T {\mathbf\Lambda},
~~~~\mathbf{K}_{\mathbf Y} = {\mathbf\Psi}^T {\mathbf\Psi},
\end{aligned}
\end{equation*}
where $\mathbf{K}_{\widetilde{\mathbf X}}$, $\mathbf{K}_{\widetilde{\mathbf Z}}$, $\mathbf{K}_{\mathbf Y} \in \mathbb{R}^{n \times n}$, and
\begin{equation*}
\begin{aligned}
&\mathbf\Phi = [\phi\left(\widetilde{\bm x}_1\right),\phi\left(\widetilde{\bm x}_2\right), \cdots, \phi\left(\widetilde{\bm x}_n\right)],\\
&\mathbf\Lambda = [\lambda\left(\widetilde{\bm z}_1\right),\lambda\left(\widetilde{\bm z}_2\right), \cdots, \lambda\left(\widetilde{\bm z}_n\right)],\\
&\mathbf\Psi = [\psi\left(\bm {y}_1\right),\psi\left({\bm y}_2\right), \cdots, \psi\left({\bm y}_n\right)].
\end{aligned}
\end{equation*}

The empirical estimation of the cross-covariance operator
$\Sigma_{\widetilde Z \widetilde X}$ can be written as
\begin{equation*}
\widehat{\Sigma}_{\widetilde Z \widetilde X}^{(n)} = \frac{\mathbf\Lambda \mathbf{H}_{n} \mathbf \Phi^T }{n},
\end{equation*}
where $\mathbf{H}_n = \mathbf{I}_n - \frac{\mathbf 1 {\mathbf 1}^T }{n}$ is the centering
matrix and $\mathbf{1} \in \mathbb{R}^{n}$  is the all-ones vector. The estimations of covariance operators $\widehat{\Sigma}_{\widetilde X \widetilde X}^{(n)}$ and
$\widehat{\Sigma}_{\widetilde Z \widetilde Z}^{(n)}$ can be derived similarly. By
regularizing the singular covariance operators
with $\varepsilon$ \cite{fukumizu2004dimensionality},
the NOCCO $V_{\widetilde Z \widetilde X}$ can be estimated by
\begin{equation*}
\widehat{V}_{\widetilde Z \widetilde X}^{(n)} = \left(\widehat{\Sigma}_{\widetilde Z \widetilde Z}^{(n)}+\varepsilon I\right)^{-1 / 2}
\widehat{\Sigma}_{\widetilde Z \widetilde X}^{(n)}\left(\widehat{\Sigma}_{\widetilde X \widetilde X}^{(n)}+\varepsilon I\right)^{-1 / 2}.
\end{equation*}

Based on the empirical estimation of the cross-covariance operators, the estimation of the COND $V_{\widetilde Z \widetilde X \mid Y}$
can be written as
\begin{equation*}
\widehat{V}_{\widetilde Z \widetilde X | Y}^{(n)} = \widehat{V}_{\widetilde Z \widetilde X}^{(n)}-\widehat{V}_{\widetilde Z Y}^{(n)} \widehat{V}_{Y \widetilde X}^{(n)}.
\end{equation*}

Let $\mathbf{G}_{\widetilde{\mathbf X}}$, $\mathbf{G}_{\widetilde{\mathbf Z}}$ and $\mathbf{G}_{\mathbf Y}$ be the centered Gram matrices, where
$\mathbf{G} = \mathbf{H}_n \mathbf{K}\mathbf{H}_n^T$.
Define $\mathbf{R}_{\widetilde{\mathbf X}}$, $\mathbf{R}_{\widetilde{\mathbf Z}}$ and
$\mathbf{R}_{\mathbf Y}$ as
$\mathbf R = \mathbf{G}{(\mathbf{G} + n \varepsilon \mathbf{I}_n )}^{-1}$,
the following proposition deduces an interpretable empirical estimation
of the conditional dependence.

\begin{theorem}
\label{Theorem:COND_empirical_estimation}
Denote $\mathbf S = \mathbf{I}_n - \mathbf{R}_{\mathbf Y}$. The empirical estimation of the conditional dependence is
\begin{align}
\label{CONDcomparison}
\hat{I}_{n}^{C O N D}(X,Z|Y) = \emph{Tr}( \mathbf{R}_{\widetilde{\mathbf Z}} \mathbf S \mathbf{R}_{\widetilde{\mathbf X}}\mathbf S).
\end{align}
\end{theorem}

Theorem~\ref{Theorem:COND_empirical_estimation} improves the results in \cite{fukumizu2007kernel} by giving a more interpretable empirical estimation of the conditional dependence measure.
It is obvious that the conditional dependence $\hat{I}_{n}^{C O N D}(X,Z|Y)$ considers all the conditions by adjusting $\mathbf{R}_{\widetilde{\mathbf Z}}$ and $\mathbf{R}_{\widetilde {\mathbf X}}$ with $\mathbf{I}_n - \mathbf{R}_{\mathbf Y}$.
The following remark about the unconditional case further explains how the condition $Y$ works in the conditional dependence.

\begin{remark} [\cite{fukumizu2007kernel}]
The empirical estimation of the dependence can be expressed as
\begin{align}\label{NOCCOcomparison}
\hat{I}_{n}^{NOCCO}(X,Z) = \emph{Tr}(\mathbf{R}_{\mathbf Z} \mathbf{I}_n  \mathbf{R}_{\mathbf X} \mathbf{I}_n ).
\end{align}
\end{remark}

Comparing Eq.~\eqref{CONDcomparison} and Eq.~\eqref{NOCCOcomparison}, we notice that there exists an interesting relationship.
The conditional information w.r.t. $\mathbf{Y}$ is reflected in $\mathbf{R}_{\mathbf Y}$, which is
used to adjust the identity weights. The intrinsic relationship between feature and domain is explored by considering all the class condition, which is vital for learning more discriminative features.

Besides, the empirical estimation in Theorem~\ref{Theorem:COND_empirical_estimation} is well defined, even the condition $Y$ is independent of $X$. More precisely, the random variable $Y$ will be a constant when the observed data are from the same class $\mathbf{y}_k$. In this case, $\hat{I}_{n}^{C O N D}(X,Z|Y)$ is still valid and will adaptively measure the unconditional dependence $\hat{I}_{n}^{NOCCO}(X,Z)$, which means that $\hat{I}_{n}^{NOCCO}(X,Z)$ is a special case of $\hat{I}_{n}^{C O N D}(X,Z|Y)$.
We conclude this property as the following corollary.

\begin{corollary}
\label{corollary: no_Y}
Assuming that $k_{\widetilde{\mathcal X}}$ and $k_{\widetilde{\mathcal Z}}$ are radial kernels. If $Y = \emptyset$ almost surely (almost everywhere), then the empirical estimations of the conditional dependence and dependence are equal.
Then,
\begin{equation*}
\label{NOCCOcomparison2}
\hat{I}_{n}^{NOCCO}(X,Z)  = \hat{I}_{n}^{COND}(X,Z|Y),
\end{equation*}
where the radial kernel is $ k(x,y)=k(\|x-y\|) $.
\end{corollary}

The condition in Corollary \ref{corollary: no_Y} reveals that $Y$ is a constant vector, which means that $Y$ is independent of both $X$ and $Z$. Therefore, we can derive $\mathbf{R}_{\mathbf{Y}} = \textit{0}$ mathematically.
Then, $\hat{I}_{n}^{COND}(X,Z|Y)$ will be equal to $\hat{I}_{n}^{NOCCO}(X,Z) $ theoretically.
Since our MCI aims to achieve the conditional independence by fully considering the influence of class-conditioned information, it is actually an improvement of the independence based methods.

\begin{theorem}[\cite{fukumizu2007kernel},Theorem 5]
\label{theorem:convergence}
Assume that $V_{ZX}$, $V_{ZY}$, and $V_{YX}$ are Hilbert-Schmidt, and that the regularization constant $\varepsilon_n$ satisfies $\varepsilon_n \rightarrow 0$ and $\varepsilon_n^3n\rightarrow \infty$, then we have
\begin{equation}
\|\widehat{V}_{XZ|Y}^{(n)} - V_{XZ|Y} \|_{HS} \rightarrow 0~~(n\rightarrow \infty)
\end{equation}
in probability with rate $\varepsilon_n^{-\frac{3}{2}}n^{-\frac{1}{2}}$.
\end{theorem}

Theorem~\ref{theorem:convergence} shows that $\widehat{V}_{XZ|Y}^{(n)}$ converges in probability to $V_{XZ|Y}$ in HS norm. In particular, the empirical measures ${\hat I}_n^{COND}$ converges to $I_n^{COND}$ at rate $\varepsilon_n^{-\frac{3}{2}}n^{-\frac{1}{2}}$. We provide the proof in
the supplementary material.

\begin{figure*}[t]
\vskip 0.27in
\begin{center}
\centering{\includegraphics[width=360pt, trim=  180 20 180 60]{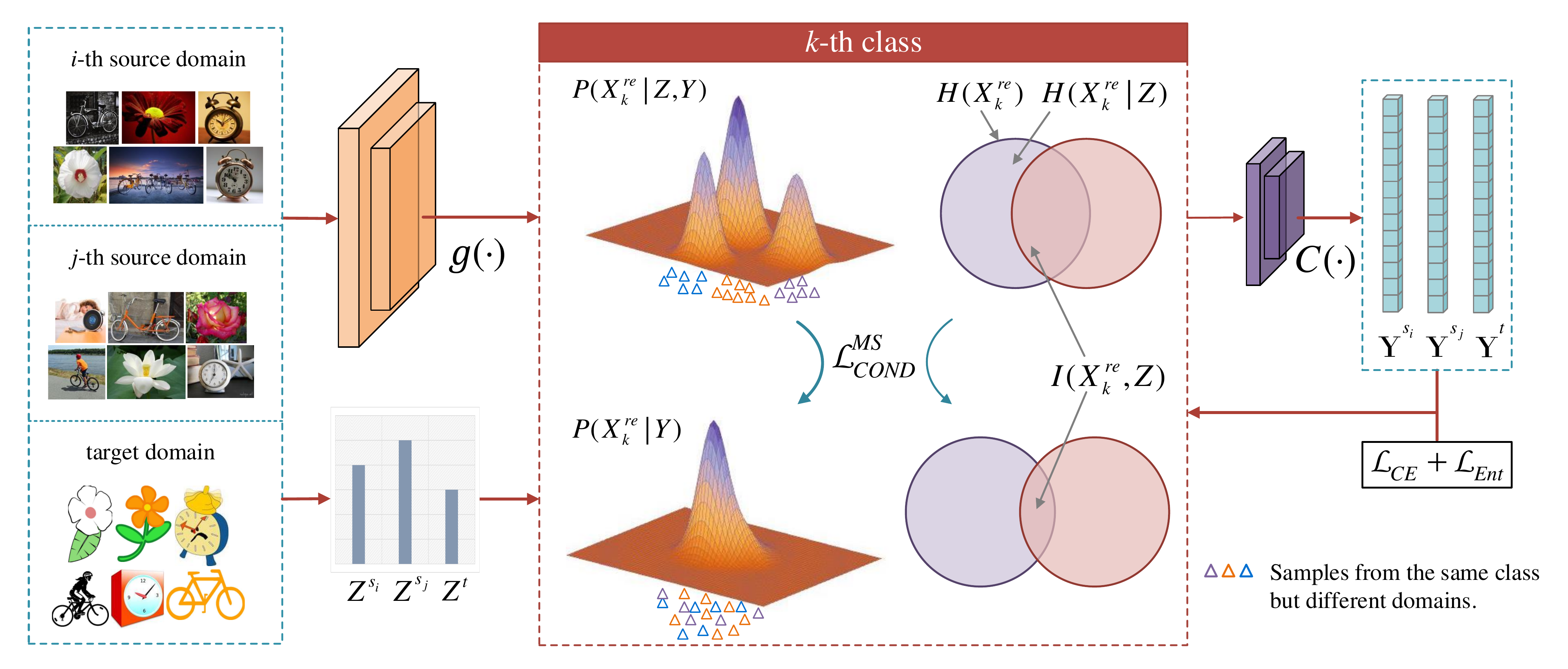}}
\caption{Flowchart of MCI/MS-MCI (described in Section~\ref{section:MCI} and Section~\ref{section:MS_MCI}).
MCI/MS-MCI explores class-conditioned domain-invariant features by achieving the statistical conditional independence.
Red arrows represents data flow of the source and target domains.
The feature transformation $g(\cdot)$ is a two layer fully connected network. The classifier $C(\cdot)$ is a single layer fully connected network.
The domain-specific information is modeled by $Z$ and the pseudo-labels of the target samples are used to construct $Y$, which make the optimization of $\mathcal{L}_{COND}^{MS}$ available.
Removing the domain-specific information by conditional independence is equal to reducing the shared information between $X_k^{re}$ and $Z$, \textit{i.e.}, $I(X_k^{re},Z)$ for each class. Based on the class-conditional independent feature $X_k^{re}$, samples from the same class tend to have an identical class-conditional distribution $P(X_k^{re}|Y)$.}
\label{fig:MCI}
\end{center}
\vskip -0.2in
\end{figure*}

\subsection{Maximizing Conditional Independence}
\label{section:MCI}
In this section, we tackle the UDA problem by proposing MCI. Guaranteed by the theoretical properties of COND, it is expected to remove the domain-specific information while preserving the class information in the feature space. The flowchart of MCI is shown in Figure~\ref{fig:MCI}.

As we aim to generalize the classifier trained on the source domain to the target domain, a supervised learning task performed on the source domain is considered.
Let $\mathbf{W}_g$ and $\mathbf{W}_C$ represent the parameters of the feature transformation $g(\cdot)$
and the classifier $C(\cdot)$, respectively.
The cross-entropy function $\mathcal{L}_{CE}$ is used to optimize the classifier with the labeled source samples, which is computed as
\begin{equation}
\label{Eq: loss CE}
\mathcal{L}_{CE}(\mathbf{W}_g,\mathbf{W}_C) = \sum_{i=1}^{K}\sum_{j=1}^{n_s} -y_{ij}^s\log \hat{y}_{ij}^s,
\end{equation}
where $\hat{y}_{ij}^s = C(g(\mathbf{x}_j^s))$ and $\sum_{i=1}^{K}\hat{y}_{ij}^s = 1$. $\hat{y}_{ij}^s$
is the prediction probability of $\mathbf{x}_j^s$ belonging to the $i\text{-th}$ class.
$y_{j}^s$ is the ground truth label of $\mathbf{x}_j$.

As mentioned in Section~\ref{Domain Private Information Removing}, we perform the class-conditioned transferring by maximizing the conditional independence, which removes the domain-specific information from the
class-conditioned feature space. As the target samples $\mathbf{X}^t$ are unlabeled, we firstly pre-train the classifier $C(\cdot)$ with loss $\mathcal{L}_{CE}$ on the source domain. Then, we initialize
and update $\hat{\mathbf{Y}}^t$ with the probability predictions from $C(\cdot)$. An intuitive illustration is presented in Figure~\ref{fig:MCI}. According to Theorem~\ref{Theorem:COND_empirical_estimation},
the conditional dependence of the feature $X$ and domain $Z$ given $Y$ is measured by the HS norm as
\begin{equation*}
\label{cond loss}
\mathcal{L}_{COND}(\mathbf{W}_g) = \hat{I}_{n}^{C O N D}(X,Z|Y).
\end{equation*}

According to Lemma \ref{Lemma: independence with operators}, minimizing loss $\mathcal{L}_{COND}$ promises to maximize the conditional independence of feature $X^{re}$ and domain $Z$ given class $Y$.
Based on the properties of the conditional independence in RKHS, samples from the same class but different domains tend to have an identical conditional distribution, and the mutual information between $X^{re}_k$ and $Z$  is close to zero. A vivid explanation is provided in Figure~\ref{fig:MCI}.
Thus, the classifier trained on the source domain will give more accurate pseudo-labels of the target samples.
The empirical estimation of the conditional dependence will be more precise and reliable then.
These two benefit from each other, which is helpful to train a more discriminative classifier.

To further facilitate the unsupervised learning on the target domain, we add the target entropy loss $\mathcal{L}_{Ent}$ into the final objective, which is formulated as
\begin{equation}
\label{Eq: loss Ent}
\mathcal{L}_{Ent}(\mathbf{W}_g,\mathbf{W}_C) = \sum_{i=1}^{K}\sum_{j=1}^{n_t} - \hat y_{ij}^t\log \hat{y}_{ij}^t,
\end{equation}
where $\hat{y}_{ij}^t = C(g(\mathbf{x}_j^t))$ and $\sum_{i=1}^{K}\hat{y}_{ij}^t = 1$. $\hat{y}_{ij}^t$ is the probability prediction of $\mathbf{x}_j^t$
belonging to the $i\text{-th}$ class.

To summarize, the objective function of our proposed method MCI consists of three parts, namely the source cross-entropy loss $\mathcal{L}_{CE}$, the conditional dependence loss $\mathcal{L}_{COND}$ and
the target entropy loss $\mathcal{L}_{Ent}$. Thus, the total loss is computed as
\begin{equation*}
\label{loss MCI}
\mathcal{L}_{MCI}(\mathbf{W}_g,\mathbf{W}_C) = \mathcal{L}_{CE} + \beta_1 \mathcal{L}_{COND} + \beta_2 \mathcal{L}_{Ent},
\end{equation*}
where $\beta_1$, $\beta_2 > 0 $ are trade-off hyper-parameters.

Following we provide a new theoretical insight based on the divergence between class-conditional distributions $P_{X|Y}^S $ and $P_{X|Y}^T$. Based on \cite{ben2010theory}, domain $\mathcal{D}=(\mu,f)$ is defined by a distribution $\mu$ on inputs $\mathcal{X}$ and a labeling function $f$. The probability according to the distribution $\mu$ that a hypothesis $h$ disagrees with a labeling function $f$ (which can also be a hypothesis) is defined as
\begin{align*}
\epsilon_{\mathcal{D}}(h) = \epsilon_{\mathcal{D}}(h,f) = \mathbb{E}_{\bm{x}\sim\mu}[\mathbb{I}(h(\bm{x}),f(\bm{x}))],
\end{align*}
where $\mathbb{I}(\cdot,\cdot)$ is an indicator function. For the source and target domains, we denote the source error and target error of a hypothesis $h$ as $\epsilon_{S}(h)$ and $\epsilon_{T}(h)$ respectively.

\begin{theorem}
\label{Theorem: single source upper bound}
Let $\mathcal{H}$ be a hypothesis space of VC dimension \textit{d}, $m$ be the sample size of source domain and $f_s$ be the ground truth labeling function for the source domain. If $\hat{h}\in \mathcal{H}$ is the empirical minimizer of $\hat{\epsilon}_S(h)$ and $h_T^* = \underset{h\in\mathcal{H}}{\textup{argmin}}~\epsilon_T(h)$ is the target error minimizer, then for any $\delta\in(0,1)$, with probability at least $1-\delta$,
\begin{align}
\label{Eq: upper bound}
\epsilon_T(\hat{h}) \leq & \epsilon_T(h^*) + 2(\lambda + \frac{1}{2} \mathbb{E}_Y[d_{\mathcal{H}\Delta\mathcal{H}}(P_{X|Y}^{S},P_{X|Y}^{T})] +  \notag \\
& \| P_Y^S - P_Y^T \|_1) + 2\eta_{d,m,\delta},
\end{align}
where $\eta_{d,m,\delta}= \sqrt{\frac{1}{2m}(log\frac{d}{\delta})}$, $\lambda = \underset{h\in\mathcal{H}}{\min}\{\epsilon_S(h) + \epsilon_T(h)\}$.
\end{theorem}

Theorem~\ref{Theorem: single source upper bound} shows the upper bound on the target error of the learned hypothesis. Here we focus on the expectation of divergence between the class-conditional distributions, \textit{i.e.},
$\mathbb{E}_Y[d_{\mathcal{H}\Delta\mathcal{H}}(\mathcal{P}_{X|Y}^{S},\mathcal{P}_{X|Y}^{T})]$, and the joint prediction error $\lambda$.
The former item evaluates the discrepancy between the class-conditional distributions between source and target domains, which motivates class-conditional distribution alignment
based methods of the class-conditional transferring.
MCI aims to remove the domain-specific information by achieving the conditional independence with $V_{\widetilde X \widetilde Z | Y}=\textit{0}$.
Besides, Theorem \ref{Theorem: conditional distribution alignment} shows that MCI derives a class-conditional distribution alignment, \textit{i.e.}, $P_{X^{re}|Y}^S = P_{X^{re}|Y}^T$, which indicates that
optimizing MCI is equal to minimizing this expectation of the class-conditioned $\mathcal{H}\Delta\mathcal{H}$-divergence.

If the joint prediction error $\lambda$ in Eq.~\ref{Eq: upper bound} is large, it is impossible to learn a classifier that performs well on both source and target domains.
Therefore, it is also important to bound $\lambda$.
Inspired by \cite{xie2018learning,zhu2020deep}, we mathematically illustrate that MCI is trying to optimize the upper bound of $\lambda$ by  utilizing the pseudo-labels. Based on the triangle inequality for classification error \cite{ben2007analysis,crammer2008learning}, \textit{i.e.}, $\epsilon(f_1,f_2) \leq \epsilon(f_1,f_3) + \epsilon(f_2, f_3)$, for any labeling functions $f_1$, $f_2$ and $f_3$, we have
\begin{align}
\lambda & =  \underset{h\in\mathcal{H}}{\min}~{\epsilon_S(h,f_s) + \epsilon_T(h,f_t)} \notag \\
&\leq \underset{h\in\mathcal{H}}{\min}~{\epsilon_S(h,f_s) + \epsilon_T(h,f_s) + \epsilon_T(f_s,f_t)}. \label{Eq: the second inequality of the joint prediction error}
\end{align}
In order to present a more clear illustration, we decompose the hypothesis into the feature transformation $g(\cdot)$ and the classifier $C(\cdot)$.
Thus, Eq.~\ref{Eq: the second inequality of the joint prediction error} can be rewritten as
\begin{align*}
 \underset{g,C}{\min}~ & \epsilon_S(C\circ g,C_s\circ g) + \epsilon_T(C \circ g,C_s \circ g) + \epsilon_T(C_s \circ g,C_t \circ g),
\end{align*}
where $f_s = C_s\circ g$ and $f_t = C_t \circ g$. The first and second items denote the disagreements between the classifier $C(\cdot)$ and the source classifier $C_s(\cdot)$ on source and target domains,
respectively.
With the supervised training on the labeled source domain, the disagreements can be decreased by approximating $C_s(\cdot)$.
The last item originally denotes the disagreement between the source labeling function $f_s$ and the target labeling function $f_t$ on the target domain, which is nonnegative.
However, if $g(\cdot)$ maps samples from the same class but different domains nearby in the latent feature space, $C_s(\cdot)$ and $C_t(\cdot)$ will have similar decision boundaries on the target domain.
It is obvious that the last item can be decreased by learning class-conditioned domain-invariant features, which can be sufficiently guaranteed by the class-conditional distribution alignment.
Thus, the three items in Eq.~\ref{Eq: the second inequality of the joint prediction error} is expected to be small. The joint prediction error $\lambda$ will be optimized by the training of MCI.

\begin{table*}[t]
    \vskip 0.018in
    \caption{Accuracies (\%) on Image-CLEF, Office-10 (AlexNet), Office-31 and Office-Home (ResNet-50).}
    \label{tab:results on 4dataset}
    \vskip 0.1in
    \renewcommand{\tabcolsep}{1.02pc}
    \renewcommand{\arraystretch}{0.6}
    \begin{center}
    \begin{small}
    \begin{tabular}{c|cccccc|c}
    \toprule
    \textbf{ImageCLEF} &  I$\rightarrow$P & P$\rightarrow$I & I$\rightarrow$C & C$\rightarrow$I & C$\rightarrow$P & P$\rightarrow$C & Mean \\
    \midrule
    Source \cite{he2016deep}             & $74.8 \pm {0.3}$ & $83.9 \pm {0.1}$ & $91.5 \pm {0.3}$ & $78.0 \pm {0.2}$ & $65.5 \pm {0.3}$ & $91.2 \pm {0.3}$ & 80.7 \\
    DAN \cite{long2018transferable}         &     $74.5 \pm {0.4}$ & $82.2 \pm {0.2}$ & $92.8 \pm {0.2}$ & $86.3 \pm {0.4}$ & $69.2 \pm {0.4}$ & $89.8 \pm {0.4}$ & 82.5 \\			
    DANN \cite{ganin2016domain}             & $75.0 \pm {0.3}$ & $86.0 \pm {0.3}$ & $96.2 \pm {0.4}$ & $87.0 \pm {0.5}$ & $74.3 \pm {0.5}$ & $91.5 \pm {0.6}$ & 85.0 \\
    CDAN+E \cite{long2018conditional}        & $77.7 \pm {0.3}$ & $90.7 \pm {0.2}$ & $97.7 \pm {0.3}$ & $91.3 \pm {0.3}$ & $74.2 \pm {0.2}$ & $94.3 \pm {0.3}$ & 87.7 \\
    KGOT \cite{zhang2019optimal} & 76.3 & 83.3& 93.5 & 87.5 & 74.8 & 89.0 & 84.1\\
    SAFN \cite{xu2019larger}                 & $78.0 \pm {0.4}$ & $91.7 \pm {0.5}$ & $96.2 \pm {0.1}$ & $91.1 \pm {0.3}$ & $77.0 \pm {0.5}$ & $94.7 \pm {0.3}$ & 88.1 \\
    ETD \cite{li2020enhanced} & 81.0$  $ & 91.7$  $ & \textbf{97.9}$  $ & 93.3$  $ & 79.5$  $ & 95.0$  $ & 89.7 \\
    DSAN \cite{zhu2020deep} & $80.2\pm{0.2}$ & $\textbf{93.3}\pm{0.4}$ & $97.2\pm{0.2}$ & $93.8\pm{0.2}$ & $80.8\pm{0.4}$ &$95.9\pm{0.4}$ & 90.2 \\
    DMP \cite{luo2020unsupervised} & $80.7 \pm {0.1}$ & $92.5 \pm {0.1}$ & $97.2 \pm {0.1}$ & $90.5 \pm {0.1}$ & $77.7 \pm {0.2}$ & $\textbf{96.2} \pm {0.2}$ & 89.1 \\
    \midrule
     MCI  & $ \textbf{82.0} \pm {0.1}$ & $ 92.8\pm {0.1}$ & $97.0 \pm {0.1}$ & $\textbf{95.8} \pm {0.1}$ & $\textbf{82.2} \pm {0.1}$ & $ 96.0 \pm {0.1}$ & \textbf{ 90.9}  \\
    \bottomrule
    \end{tabular}
    \vskip 0.1in
    \renewcommand{\tabcolsep}{0.5pc}
    \renewcommand{\arraystretch}{0.6}
    \begin{tabular}{c|cccccccccccc|c}
    \toprule
    \textbf{Office-10} & A$\to$C & A$\to$D & A$\to$W & C$\to$A & C$\to$D & C$\to$W &
            D$\to$A & D$\to$C & D$\to$W & W$\to$A & W$\to$C & W$\to$D & Mean \\
    \midrule
    Source \cite{krizhevsky2017imagenet} & 82.7 & 85.4 & 78.3 & 91.5 & 88.5 & 83.1 & 80.6 & 74.6 & 99.0 & 77.0 & 69.6 & 100.0 & 84.2 \\
    GFK \cite{gong2012geodesic} & 78.1 & 84.7 & 76.3 & 89.1 & 88.5 & 80.3 & 89.0 & 78.4 & 99.3 & 83.9 & 76.2 & 100.0 & 85.3 \\
    CORAL \cite{sun2016return} & 85.3 & 80.8 & 76.3 & 91.1 & 86.6 & 81.1 & 88.7 & 80.4 & 99.3 & 82.1 & 78.7 & 100.0 & 85.9 \\
    OT-IT \cite{courty2016optimal} & 83.3 & 84.1 & 77.3 & 88.7 & 90.5 & 88.5 & 83.3 & 84.0 & 98.3 & 88.9 & 79.1 & 99.4 & 87.1 \\	
    KGOT \cite{zhang2019optimal} & 85.7 & 86.6 & 82.4 & 91.4 & 92.4 & 87.1 & 91.8 & 85.6 & 99.3 & 89.7 & 85.0 & 100.0 & 89.7 \\
    DMP \cite{luo2020unsupervised} & 86.6 & 90.4 & 91.3 & 92.8 & 93.0 & 88.5 & 91.4 & 85.3 & 97.7 & 91.9 & 85.6 & 100.0 & 91.2 \\
    \midrule
    MCI & \textbf{87.9} & \textbf{92.7} & \textbf{96.0} &\textbf{93.7} & \textbf{94.5}& \textbf{94.1} &\textbf{93.1} &\textbf{ 87.0 }& \textbf{99.6}& \textbf{93.8} & \textbf{86.6} & \textbf{100.0} & \textbf{93.2} \\
    \bottomrule
    \end{tabular}
    \end{small}
    \end{center}
    \vspace{-2pt}
    \renewcommand{\tabcolsep}{1.0pc}
    \renewcommand{\arraystretch}{0.6}
    \begin{center}
    \begin{small}
    \begin{tabular}{c|cccccc|c}
    \toprule
    \textbf{Office-31}  & A$\rightarrow$W & D$\rightarrow$W & W$\rightarrow$D & A$\rightarrow$D & D$\rightarrow$A & W$\rightarrow$A & Mean \\
    \midrule
    Source \cite{he2016deep}             & $68.4 \pm {0.2}$ & $96.7 \pm {0.1}$ & $99.3 \pm {0.1}$ & $ 68.9 \pm {0.2}$ & $62.5 \pm {0.3}$ & $60.7 \pm { 0.3}$ & 76.1 \\
    DAN \cite{long2018transferable}             & $80.5 \pm {0.4}$ & $97.1 \pm {0.2}$ & $99.6 \pm {0.1}$ & $ 78.6 \pm {0.2}$ & $63.6 \pm {0.3}$ & $62.8 \pm { 0.2}$ & 80.4 \\				
    DANN \cite{ganin2016domain}             & $82.0 \pm {0.4}$ & $96.9 \pm {0.2}$ & $99.1 \pm {0.1}$ & $ 79.7 \pm {0.4}$ & $68.2 \pm {0.4}$ & $67.4 \pm { 0.5}$ & 82.2 \\
    CDAN+E \cite{long2018conditional}       & $94.1 \pm {0.1}$ & $98.6 \pm {0.1}$ & $\textbf{100.0} \pm {0.0}$ & $\textbf{92.9} \pm {0.2}$ & $71.0 \pm {0.3}$ & $69.3 \pm {0.3}$ & 87.7 \\
    KGOT \cite{zhang2019optimal}& 75.3 & 96.2 & 98.4 & 80.3 & 65.2 & 63.5 & 79.8 \\
    SAFN \cite{xu2019larger}                & $88.8 \pm {0.4}$ & $98.4 \pm {0.0}$ & $99.8 \pm {0.0}$ & $ 87.7 \pm {1.3}$ & $69.8 \pm {0.4}$ & $69.7 \pm { 0.2}$ & 85.7 \\
    ETD \cite{li2020enhanced} & 92.1$  $ & \textbf{100.0}$  $ & \textbf{100.0}$  $ & 88.0$  $ & 71.0$  $ & 67.8$  $ & 86.2 \\
    DSAN \cite{zhu2020deep} & $\textbf{93.6} \pm {0.2}$ & $98.3 \pm{0.1}$ &$\textbf{100.0}\pm{0.0}$&$90.2\pm{0.7}$&$73.5\pm{0.5}$& $\textbf{74.8}\pm{0.4}$ &88.4\\
    DMP \cite{luo2020unsupervised} & $93.0 \pm {0.3}$ & $99.0 \pm {0.1}$ & $\textbf{100.0} \pm {0.0}$ & $91.0 \pm {0.4}$ & $71.4 \pm {0.2}$ & $70.2 \pm {0.2}$ & 87.4 \\
    \midrule
    MCI  & $92.5 \pm {0.2}$ & $98.7 \pm {0.0}$ & $\textbf{100.0} \pm {0.0}$ & $ 92.4 \pm {0.3}$ & $\textbf{75.4} \pm {0.1}$ & $74.2 \pm {0.1}$ & \textbf{88.9}  \\
    \bottomrule
    \end{tabular}
    \vskip 0.1in
    \renewcommand{\tabcolsep}{0.3pc}
    \renewcommand{\arraystretch}{0.6}
    \begin{tabular}{c|cccccccccccc|c}
    \toprule
    \textbf{Office-Home} & Ar$\to$Cl & Ar$\to$Pr & Ar$\to$Rw & Cl$\to$Ar & Cl$\to$Pr & Cl$\to$Rw &
    Pr$\to$Ar & Pr$\to$Cl & Pr$\to$Rw & Rw$\to$Ar & Rw$\to$Cl & Rw$\to$Pr & Mean \\
    \midrule
    Source \cite{he2016deep} & 34.9 & 50.0 & 58.0 & 37.4 & 41.9 & 46.2 & 38.5 & 31.2 & 60.4 & 53.9 & 41.2 & 59.9 & 46.1 \\
    DAN \cite{long2018transferable} & 43.6 & 57.0 & 67.9 & 45.8 & 56.5 & 60.4 & 44.0 & 43.6 & 67.7 & 63.1 & 51.5 & 74.3 & 56.3 \\				
    DANN \cite{ganin2016domain} & 45.6 & 59.3 & 70.1 & 47.0 & 58.5 & 60.9 & 46.1 & 43.7 & 68.5 & 63.2 & 51.8 & 76.8 & 57.6 \\
    CDAN+E \cite{long2018conditional} & 50.7 & 70.6 & 76.0 & 57.6 & 70.0 & 70.0 & 57.4 & 50.9 & 77.3 & 70.9 & 56.7 & 81.6 & 65.8 \\
    KGOT \cite{zhang2019optimal}&36.2&59.4&65.0&48.6&56.5&60.2&52.1&37.8&67.1&59.0&41.9&72.0&54.7\\
    SAFN \cite{xu2019larger} & 52.0 & 71.7 & 76.3 & 64.2 & 69.9 & 71.9 & 63.7 & 51.4 & 77.1 & 70.9 & 57.1 & 81.5 & 67.3 \\
    ETD \cite{li2020enhanced} & 51.3 & 71.9 & \textbf{85.7} & 57.6 & 69.2 & 73.7 &  57.8 & 51.2 & 79.3 & 70.2 & 57.5 & 82.1 & 67.3\\
    DSAN \cite{zhu2020deep} & \textbf{54.4} & 70.8 & 75.4 & 60.4 & 67.8 & 68.0 & 62.6 & \textbf{55.9} & 78.5 & \textbf{73.8} & \textbf{60.6} & 83.1 & 67.6 \\
    DMP \cite{luo2020unsupervised} & 52.3 & 73.0 & 77.3 & \textbf{64.3} & 72.0 & 71.8 & 63.6 & 52.7 & 78.5 & 72.0 & 57.7 & 81.6 & 68.1 \\
    \midrule
    MCI & 51.7 & \textbf{76.3} & 80.1 &60.6 & \textbf{75.2} & \textbf{76.3} & \textbf{64.8} & 51.4 & \textbf{81.7} & 69.3 & 54.8& \textbf{83.3} & \textbf{68.8} \\
    \bottomrule
    \end{tabular}

    \end{small}
    \end{center}
    \vskip -0.2in
 \end{table*}

\subsection{Extension for Multi-source Domain Adaptation}
\label{section:MS_MCI}
MCI explores the conditional independence between feature $X$ and domain $Z$ given class $Y$, which takes the domain-specific information into consideration. Consequently, it's natural and elegant to extend MCI
for the MDA scenario.

Combing multiple source domains into a single-source domain, and then we can apply MCI directly to solve the MDA problem. However, this straightforward way is inappropriate,
since it ignores the discrepancy across source domains brought by different domain-specific information.
In this paper, we propose multi-source MCI (MS-MCI), which treats each source separately. The framework of MS-MCI is presented in Figure~\ref{fig:MCI}.

Denote multiple similar but not identical source domains as
$\{\mathcal{D}^{S_1}, \mathcal{D}^{S_2}, \cdots, \mathcal{D}^{S_N} \}$, where $N$ is the number of source domains. The objective of MS-MCI can be formulated as
\begin{align*}
\label{Eq: objective of MS-MCI}
\mathcal{L}_{MS-MCI}(\mathbf{W}_g, \mathbf{W}_C) = \sum_{i = 1}^N \mathcal{L}_{CE}^i + \beta_1 \mathcal{L}_{COND}^{MS} + \beta_2 \mathcal{L}_{Ent},
\end{align*}
where loss $\mathcal{L}_{CE}^{i}$ is the cross-entropy loss $\mathcal{L}_{CE}$ (Eq.~\ref{Eq: loss CE}) for the $i$-th source domain, and the target entropy loss $\mathcal{L}_{Ent}$ keeps the same as Eq.~\ref{Eq: loss Ent}.
The conditional dependence loss $\mathcal{L}_{COND}^{MS}$ is extended from $\mathcal{L}_{COND}$.
Specifically, the sample space of $Z$ under multi-source scenario changes to $\{\mathbf{z}^{s_1},\cdots, \mathbf{z}^{s_N}, \mathbf{z}^{t}\}$.
In order to learn class-conditioned domain-invariant features $\mathbf{X}^{re}$, we construct feature matrix $\mathbf{X}\in\mathbb{R}^{d\times n}$ and domain label matrix $\mathbf{Z}\in \mathbb{R}^{(N+1)\times n}$
based on samples from all source and target domains, where $n = \sum_{i = 1}^N n_{s_i} + n_t $. Pseudo-labels of target samples are also employed in $\mathbf{Y}\in \mathbb{R}^{K\times n}$.
And then the computation of $\mathcal{L}_{COND}^{MS}$ also follows Eq.~\ref{CONDcomparison} in Theorem~\ref{Theorem:COND_empirical_estimation}.

Different from some MDA methods that have private classifier for each source domain, MS-MCI shares both the feature transformation $g(\cdot)$
and the classifier $C(\cdot)$ across all domains. Thus, the framework of MS-MCI can be easier to apply.
Similar to the optimization of MCI, MS-MCI is also firstly trained with the cross-entropy loss $\sum_{i = 1}^N \mathcal{L}_{CE}^i$ on all source domains for giving
more reliable pseudo-labels for target samples.
The variousness of training data avoids overfitting in the supervised learning,
which leads a better performance in predicting the label of target samples.
Thus, the empirical estimation of $\mathcal{L}_{COND}^{MS}$ is supposed to be more reliable.

Following we derive an upper bound based on the $\mathcal{H}\Delta\mathcal{H}$-divergence between the class-conditional distributions for the multi-source scenario.

\begin{theorem}
\label{Theorem: Multi-source domain bounds}
Let $\mathcal{H}$ be a hypothesis space of VC dimension \textit{d}. Let $m$ be the size of labeled samples from each source domain $\{\mathcal{D}^{S_i}\}_{i = 1}^N$, $S_i$ be the labeled sample set of size $\alpha_i m$ ($\sum_i \alpha_i = 1, \alpha_i \in (0,1]$) drawn from the distribution of the $i$-th source domain, and labeled by the ground truth labeling function $f_{s_i}$. If $\hat{h} \in \mathcal{H}$ is the empirical minimizer of $\hat{\epsilon}_{\bm \alpha}(h)$ for the weight vector $\bm \alpha$ and $h_T^* = \underset{h\in\mathcal{H}}{\textup{argmin}}~\epsilon_T(h)$ is the target error minimizer, then for any $\delta \in (0,1)$, with probability at least $1- \delta$,
\begin{align*}
\epsilon_T(\hat{h}) \leq & \epsilon_T(h^*) + 2\lambda_{\bm \alpha}+ \sum_{i = 1}^N2{\alpha_i}(\frac{1}{2}\mathbb{E}_Y[d_{\mathcal{H}\Delta\mathcal{H}}(P_{X|Y}^{S_i},P_{X|Y}^{T})] + \\
& \| P_Y^{S_i} - P_Y^T \|_1) + 2\eta_{d,m,\delta},
\end{align*}
where $\eta_{d,m,\delta}= \sqrt{\frac{1}{2m}( log\frac{d}{\delta})}$ and $\lambda_{\bm \alpha} = \underset{h\in\mathcal{H}}{\min}\{\sum_{i = 1}^{N}\alpha_i\epsilon_{S_i}(h) + \epsilon_T(h)\}$.
\end{theorem}

Theorem~\ref{Theorem: Multi-source domain bounds} shows the upper bound on the target error of the learned hypothesis for the multi-source scenario. Here we also focus on the expectation of divergence
$\mathbb{E}_Y[d_{\mathcal{H}\Delta\mathcal{H}}(P_{X|Y}^{S_i},P_{X|Y}^{T})]$ and joint prediction error $\lambda_{\bm{\alpha}}$.
Instead of aligning the marginal distributions of each source and target domains, the former item motivates a fine-grained class-conditional distribution alignment across domains.
Since MS-MCI tries to guarantee that samples from the same class but different domains have identical class-conditional distributions, it is equal to optimizing this divergence item in some way.
Based on the definition of the $\mathcal{H}\Delta\mathcal{H}$-divergence \cite{ben2010theory}, it is easy to derive the following triangle inequality
\begin{align*}
d_{\mathcal{H}\Delta\mathcal{H}}(P_{X|Y}^{S_1},P_{X|Y}^{S_2}) \leq \sum_{i = 1}^2 d_{\mathcal{H}\Delta\mathcal{H}}(P_{X|Y}^{S_i},P_{X|Y}^{T}).
\end{align*}
This inequality implies that the class-conditional distribution divergence between each pair of source domains is the lower bound of the divergence between the source and target domains.
Therefore, aligning the class-conditional distributions for each pair of source domains is necessary. Since MS-MCI takes all the domain-specific information into account, it also aligns the class-conditional distribution between each pair of source domains potentially. Similar to our analysis of the joint prediction error $\lambda$ in the
single-source scenario, different domains are motivated by MS-MCI to have similar decision boundaries in the latent feature space. Thus, the joint prediction error $\lambda_{\bm \alpha}$ is expected to
be decreased by the optimization of our MS-MCI.

\begin{table*}
 \setlength{\abovecaptionskip}{0cm}
 \setlength{\belowcaptionskip}{-1cm}
\caption{Accuracies (\%) of ablation study.}
 \label{tab:ablation_experiment}
 \renewcommand{\tabcolsep}{0.5pc}
 \renewcommand{\arraystretch}{0.6}
 \begin{center}
 \begin{small}
 \begin{tabular}{m{3.0cm}<{\centering}|m{0.7cm}<{\centering}m{0.7cm}<{\centering}m{0.7cm}<{\centering}|m{0.7cm}<{\centering}m{0.7cm}<{\centering}m{0.7cm}<{\centering}|m{1.3cm}<{\centering}m{1.3cm}<{\centering}m{1.3cm}<{\centering}m{1.3cm}<{\centering}}
 \toprule
 \multirow{2}{*}{Method} &  \multicolumn{3}{c|} {Image-CLEF} &\multicolumn{3}{c|}{Office-31}& \multicolumn{4}{c}{Office-Home}\\
 &C$\rightarrow$P & I$\rightarrow$P & C$\rightarrow$I & A$\rightarrow$D & A$\rightarrow$W & W$\rightarrow$A & Cl$\rightarrow$ Rw & Pr $\rightarrow$ Ar & Rw $\rightarrow$ Ar & Rw $\rightarrow$ Pr \\
 \midrule
 MCI (w/o $\mathcal{L}_{COND}$)  & 77.6& 79.5 & 91.5 & 89.9 & 89.7 & 70.3 & 70.3 & 61.1 & 67.6 & 80.8 \\
 MCI (w/o $\mathcal{L}_{Ent}$)   & 81.8& 81.7 & 95.5 & 91.6 & 91.3 & 74.1 & 75.5 & 64.4 & 69.1 & 83.0 \\
 HSIC                           & 80.7& 81.6 & 95.1 & 87.3 & 90.2 & 72.7 & 71.3 & 61.6 & 68.0 & 82.2 \\
 MCI                            & \textbf{82.2}& \textbf{82.0} & \textbf{95.8} & \textbf{92.4} & \textbf{92.5} &\textbf{74.2} & \textbf{76.3} & \textbf{64.8} & \textbf{69.3} & \textbf{83.3}\\
 \bottomrule
 \end{tabular}
 \end{small}
 \end{center}
 \vskip -0.1in
\end{table*}

\section{Experiments}
\label{Experiments}
In this section, MCI and MS-MCI are both evaluated and compared with state of the art domain adaptation methods on four standard visual benchmarks.

\textbf{Image-CLEF} \cite{caputo2014imageclef} has 3 domains with 12 classes, \textit{i.e.}, \textit{Caltech} (C), \textit{ImageNet} (I), \textit{Pascal} (P). Especially, it is a balanced dataset as each domain contains 600 images. For multi-source scenario, we choose one domain as the target domain and others as the source domains in turn, which is the same for the following datasets.

\textbf{Office-31} \cite{saenko2010adapting}
consists of 3 domains with 31 classes, \textit{i.e.},
\textit{Amazon} (A) (images downloaded from online merchants), \textit{Webcam} (W) (low-resolution images by a web camera), \textit{DSLR} (D) (high-resolution images from a digital SLR camera). This dataset not only captures a large intra-class variation, but also represents several different visual dataset shifts.

\textbf{Office-10} \cite{gong2012geodesic} contains 2533 images from 4 domains with 10 classes, \textit{i.e.}, \textit{Amazon} (A), \textit{Caltech} (C), \textit{DSLR} (D), and \textit{Webcam} (W). Compared with Office-31, the \textit{Caltech} domain constructed from Caltech-256 \cite{griffin2007caltech} is added as the fourth domain.

\textbf{Office-Home} \cite{venkateswara2017deep}
is a medium-sized dataset, which consists 15500 images from 4 domains with 65 classes. The domains include:
\textit{Artistic} (Ar), \textit{Clipart} (Cl), \textit{Product} (Pr), and \textit{Real-World} (Rw). Each class has around 70 images and 99 images maximally.

\subsection{Numerical Implementation}
Deep neural network based methods have achieved considerable performance
in UDA. Thus, we construct the feature matrices $\mathbf{X}^s$ and $\mathbf{X}^t$ of the source and target domains
by utilizing the AlexNet features \cite{wang2020unsupervised} for Office-10 and ResNet-50 features \cite{wang2020unsupervised} for others.
The feature transformation $g(\cdot)$ is a two layer fully connected network with 512
output units. The classifier $C(\cdot)$ is a single fully connected layer with $K$ output
units and a softmax activate function. All the kernels appeared in this paper are Gausian kernel
$k(\mathbf{x},\mathbf{x}^{\prime}) = \text{exp}(-\|\mathbf{x}-\mathbf{x}^{\prime}\|_2^2/{\sigma}^2)$
for the requirement of characteristic property, where $\sigma^2$ equals to the mean of
all the square Euclidean distances $\|\mathbf{x}-\mathbf{x}^{\prime}\|_2^2$.

We firstly pre-train the classifier $C(\cdot)$ with loss $\mathcal{L}_{CE} $ on the source domain. $\hat{\mathbf{Y}}^t$ is initialized with the probability predictions of
the pre-trained $C(\cdot)$. Then, the whole model will be trained with the total loss $\mathcal{L}_{MCI}$. In the iterative training process, $\hat{\mathbf{Y}}^t$ will be updated every epoch.
Finally, MCI will achieve the class-conditioned transferring and narrow the distance between the conditional distributions by minimizing the conditional dependence between the feature $X$ and the domain $Z$. The training process for MS-MCI is similar to MCI.

The whole algorithm is implemented with the deep learning framwork PyTorch \cite{paszke2019pytorch}. For optimization, we use the Adam optimizer through back-propagation.
The training progress is efficiency as the networks are relatively shallow compared with other deep UDA methods. We report the average classification accuracy and standard error of ten random trials.

\subsection{Results and analysis for MCI}
\textbf{Results}. The classification accuracies on Image-CLEF are shown in the top of Table~\ref{tab:results on 4dataset}.
MCI substantially outperforms all the advanced comparison methods on most
transfer tasks.
MCI performs the best on average with accuracy 90.9\%.
Compared with the second best model DSAN \cite{zhu2020deep}, the accuracy of our MCI increases by 1.4\%
and 2.0\% on tasks C$\rightarrow$P and C$\rightarrow$I, respectively.

The classification accuracies on Office-10 are shown in the second row of Table~\ref{tab:results on 4dataset}.
It is observed that MCI outperforms the other methods with a large margin on all tasks.
Especially, MCI achieves the highest mean accuracy 93.2\%, which is improved by 2.0\% compared with the second best model
DMP \cite{luo2020unsupervised}. This indicates that maximizing the conditional independence is effective in learning discriminative features.

\begin{figure}[t]
\vskip -0.1in
\subfigure[Task I $\rightarrow$ P]{\label{fig:parm_P2I}
    \begin{minipage}[b]{0.47\linewidth}
    \centering{
    \includegraphics[width=1\textwidth,trim= 20 195 10 180]{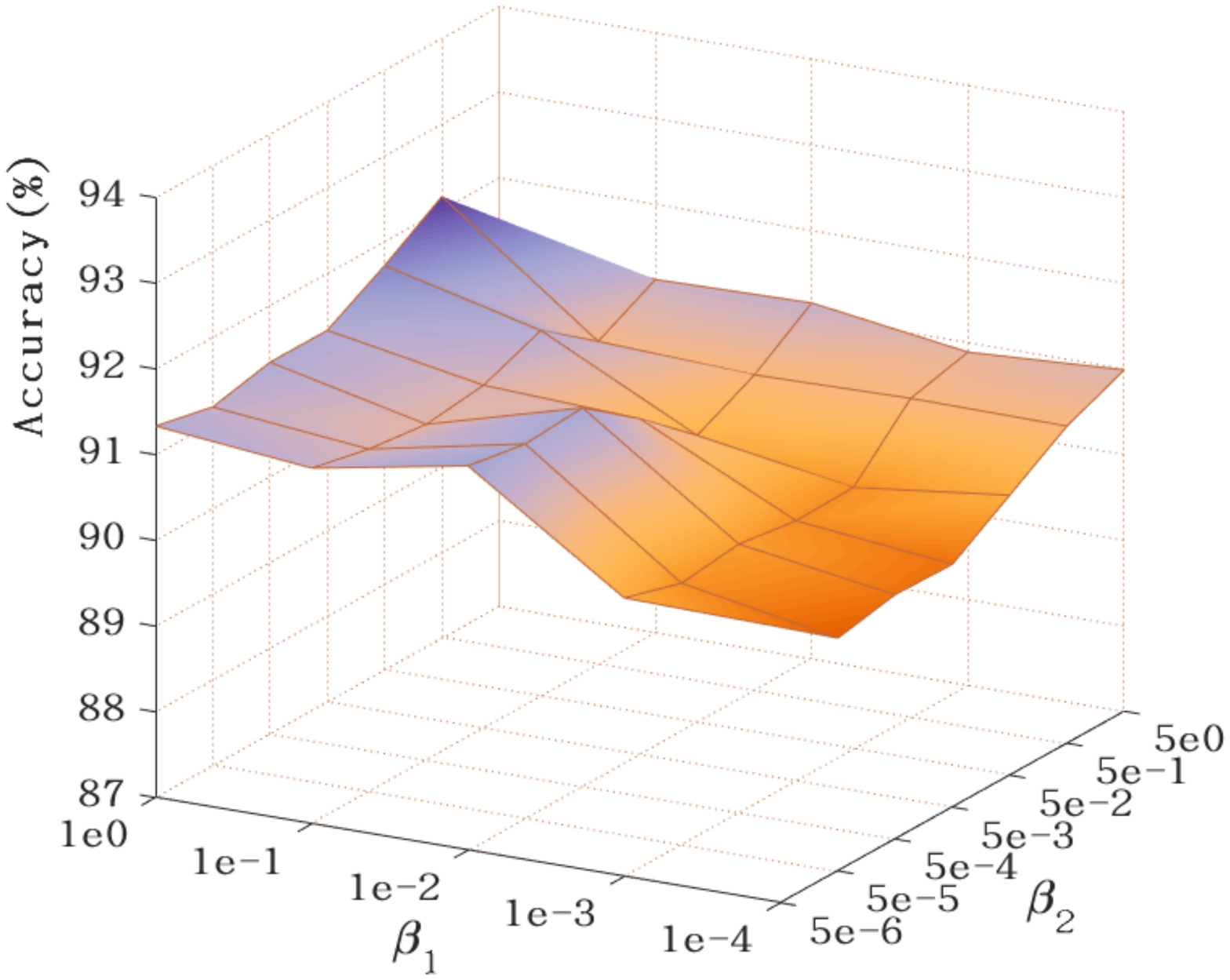}}
    \end{minipage}}
\subfigure[Task P $\rightarrow$ I]{\label{fig:parm_P2C}
    \begin{minipage}[b]{0.47\linewidth}
    \centering{
    \includegraphics[width=1\textwidth,trim= 20 180 0 190]{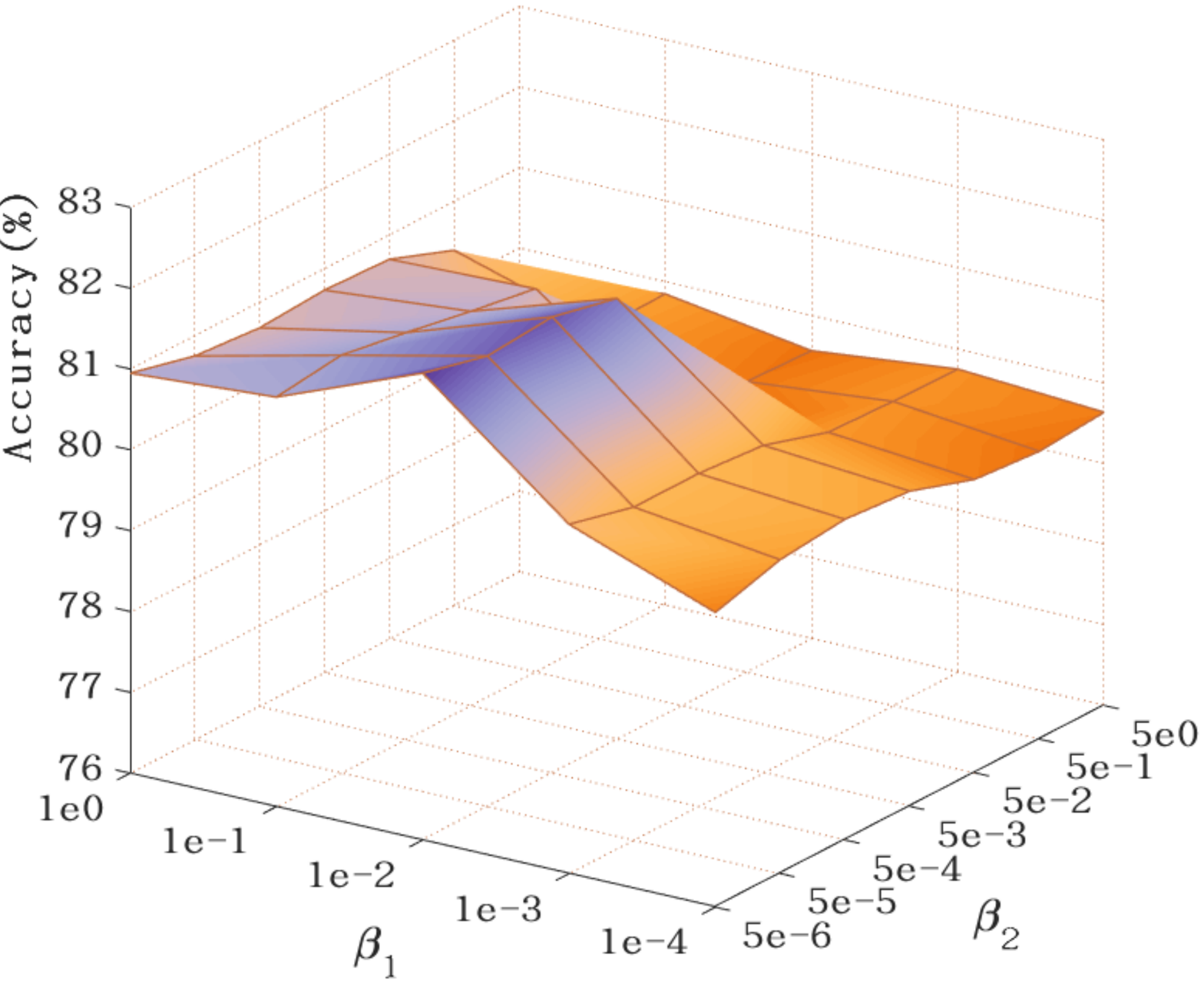}}
    \end{minipage}}
\caption{Parameter sensitivity of $\beta_1$ and $\beta_2$
on Image-CLEF tasks.}
\label{fig:parameters}
\vskip -0.1in
\end{figure}
\begin{figure*}[t]
\vskip 0.03in
\subfigure[ Before adaptation]{\label{fig:office-31-before-domain}
    \begin{minipage}[b]{0.23\linewidth}
    \centering{
    \includegraphics[width=1\textwidth,trim=114 300 114 300]{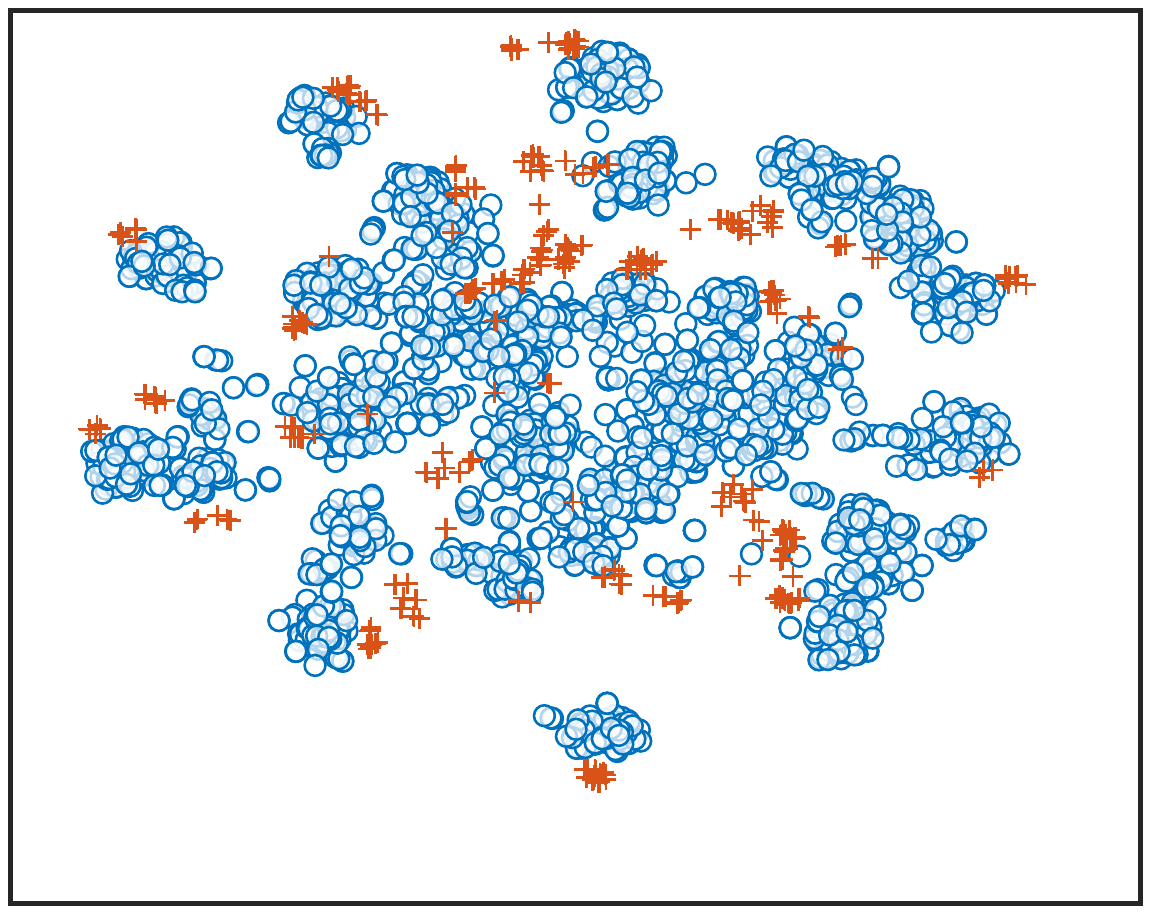}}
    \end{minipage}}
\subfigure[ After adaptation]{\label{fig:office-31-after-domain}
    \begin{minipage}[b]{0.23\linewidth}
    \centering{
    \includegraphics[width=1\textwidth,trim=114 300 114 300]{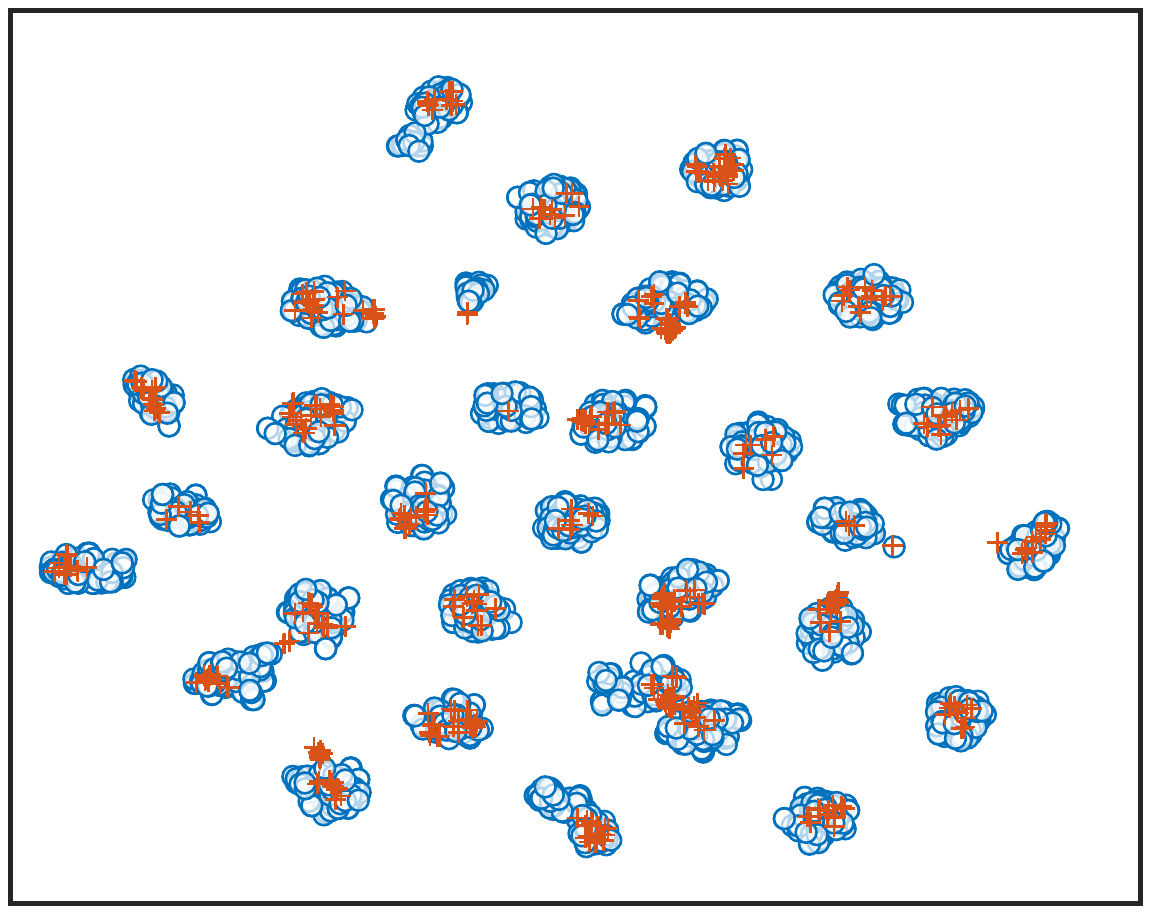}}
    \end{minipage}}
\subfigure[ Before adaptation]{\label{fig:office-31-before-class}
    \begin{minipage}[b]{0.23\linewidth}
    \centering{
    \includegraphics[width=1\textwidth,trim=114 300 114 300]{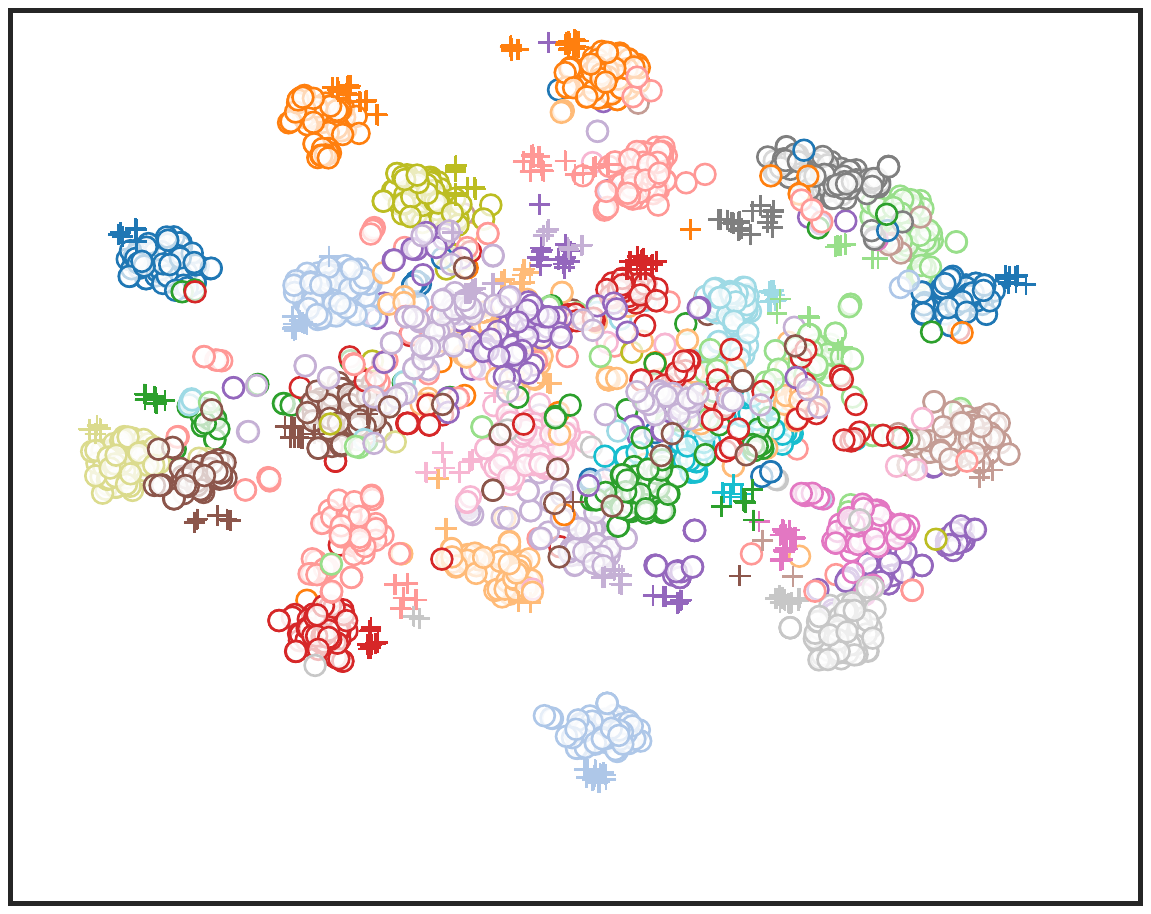}}
    \end{minipage}}
\subfigure[ After adaptation]{\label{fig:office-31-after-class}
    \begin{minipage}[b]{0.23\linewidth}
    \centering{
    \includegraphics[width=1\textwidth,trim=114 300 114 300]{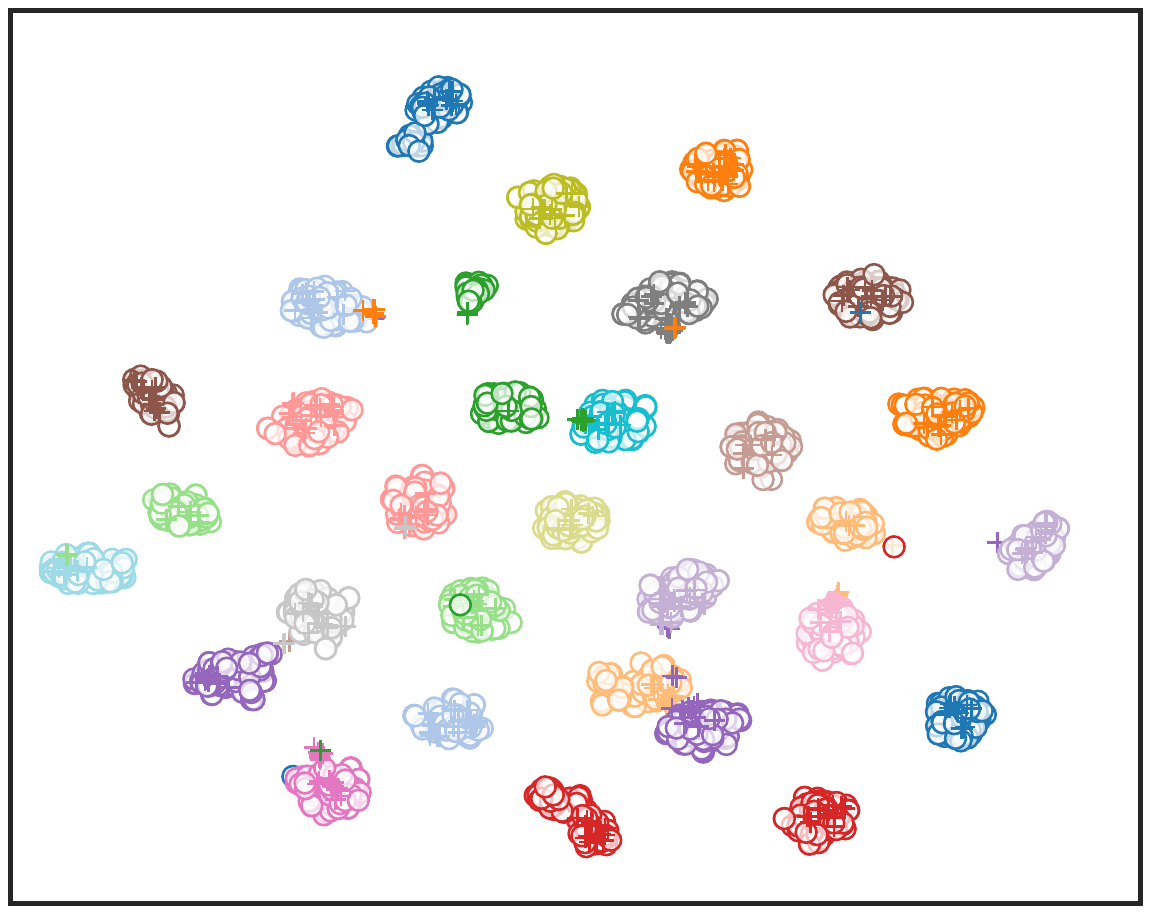}}
    \end{minipage}}
\caption{The t-SNE features are generated by Source and MCI models on Office-31 task A$\rightarrow$D, respectively.
Note: ``o'' means source domain A and ``+'' means target domain D. (a) and (b) are colored by domain. (c) and (d) are colored
by class. Best viewed in color.}\label{fig:tsne}
\end{figure*}

\begin{figure*}[t]
\subfigure[MFSAN]{\label{fig:MFSAN-office-31-before-domain}
    \begin{minipage}[b]{0.23\linewidth}
    \centering{
    \includegraphics[width=1\textwidth,trim=114 300 114 300]{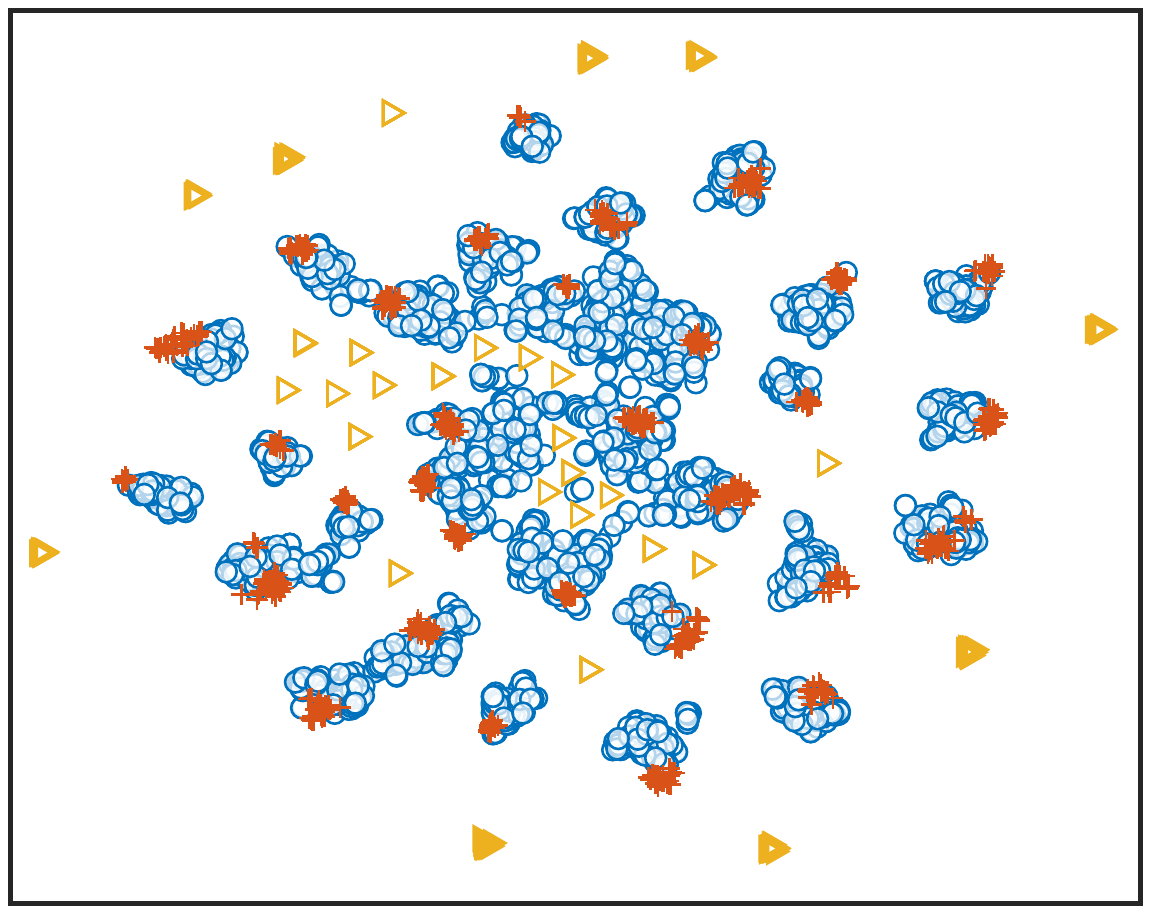}}
    \end{minipage}}
\subfigure[MS-MCI]{\label{fig:MFSAN-office-31-after-domain}
    \begin{minipage}[b]{0.23\linewidth}
    \centering{
    \includegraphics[width=1\textwidth,trim=114 300 114 300]{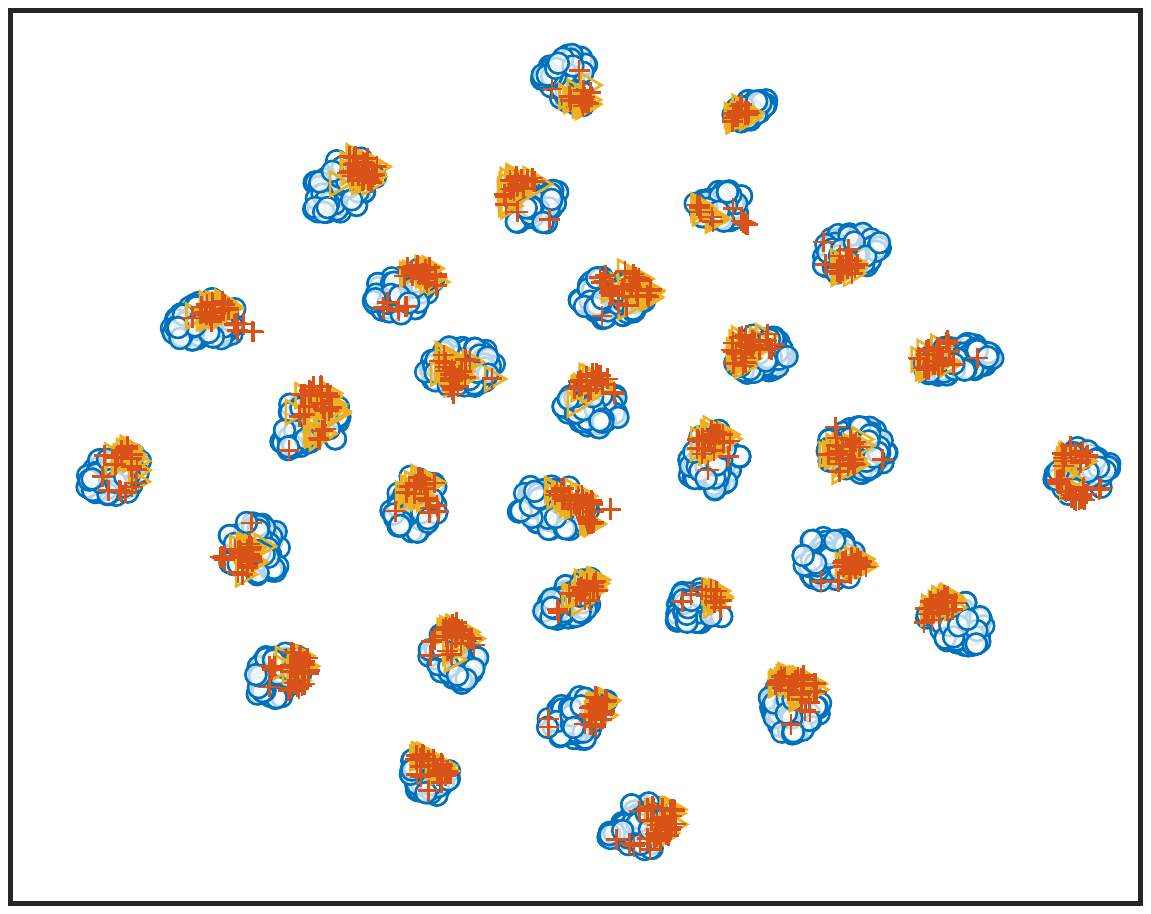}}
    \end{minipage}}
\subfigure[MFSAN]{\label{fig:MS-MCI-office-31-before-class}
    \begin{minipage}[b]{0.23\linewidth}
    \centering{
    \includegraphics[width=1\textwidth,trim=114 300 114 300]{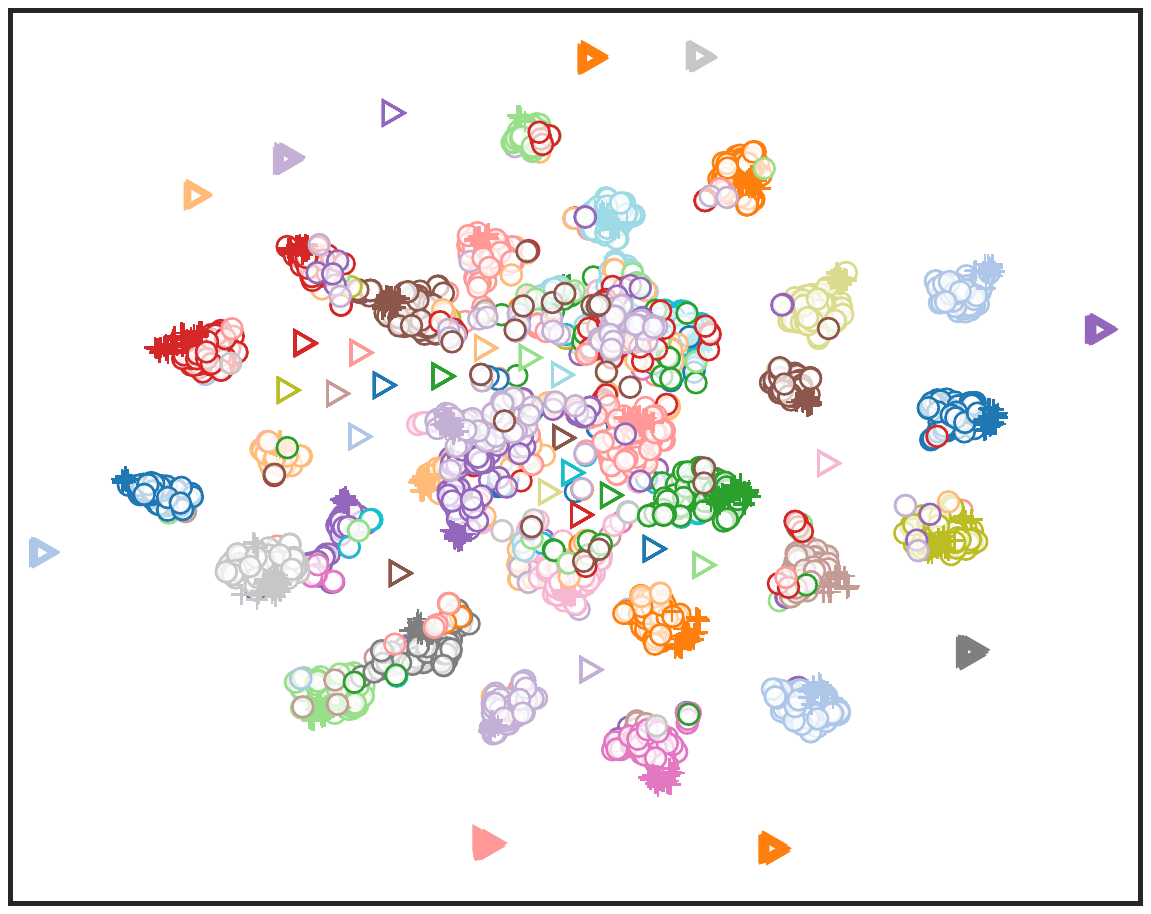}}
    \end{minipage}}
\subfigure[MS-MCI]{\label{fig:MS-MCI-office-31-after-class}
    \begin{minipage}[b]{0.23\linewidth}
    \centering{
    \includegraphics[width=1\textwidth,trim=114 300 114 300]{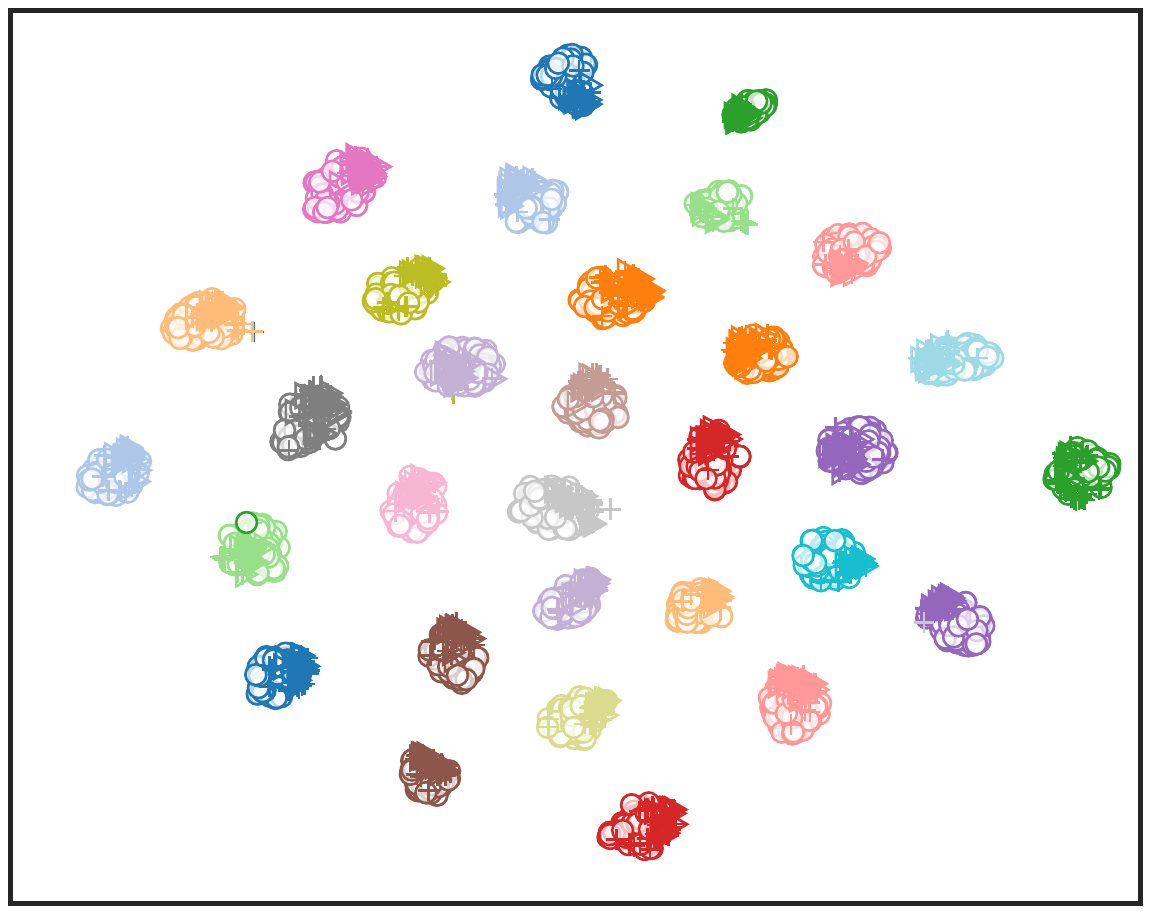}}
    \end{minipage}}
\caption{The t-SNE features are generated by MFSAN and MS-MCI models on Office-31 task A,D$\rightarrow$W, respectively.
Note: ``o'' means source domain A, ``$\triangle$'' means source domain D and ``+'' means target domain A. (a) and (b) are colored by domain. (c) and (d) are colored
by class. Best viewed in color.}
\label{fig:tsne-MS-MCI}
\end{figure*}

The classification accuracies on Office-31 are shown in the third row of Table~\ref{tab:results on 4dataset}.
MCI achieves the best average accuracy with 88.9\%.
It is worth noting that MCI achieves the highest accuracies on the hard tasks A$\rightarrow$D and D$\rightarrow$A, where A and D have a larger visual dataset shift than other tasks.
Results of MCI on other tasks are slightly lower than the best but have basically reached the state-of-the-art level.

The classification accuracies on Office-Home are shown in the bottom of Table~\ref{tab:results on 4dataset}.
Office-Home has a large intra-class variation and visual dataset shifts due to 65 classes.
Therefore, methods based on the marginal distribution alignment will suffer a serious misalignment of samples from the same class but different domains.
The class-conditioned transferring method MCI exceeds the latest methods with the average accuracy 68.8\%.
It is  also reasonable that MCI performs much better than the optimal transport based methods
KGOT \cite{zhang2019optimal} and ETD \cite{li2020enhanced}.

Comparing MCI with the most recent subdomain adaptation method DSAN \cite{zhu2020deep},
the accuracies of MCI are even 8.3\% and 7.4\% higher on the tasks Cl$\rightarrow$Rw and Cl$\rightarrow$Pr, respectively.
The encouraging results indicate that removing the domain-specific information by maximizing the conditional independence has
significant advantages in the class-conditioned transferring.

\textbf{Parameter Sensitivity}. There are two important parameters in MCI, where $\beta_1$ acts on the loss $\mathcal{L}_{COND}$ and
$\beta_2$ is to balance the target entropy loss $\mathcal{L}_{Ent}$.
Empirically, the regularization parameter $\varepsilon$ is provided sufficiently small.
For Image-CLEF and Office-10, we fix $\varepsilon=10^{-5}$.
For Office-31 and Office-Home, we fix $\varepsilon= 10^{-4}$.

We evaluate the parameter sensitivity of $\beta_1$ and $\beta_2$ on Image-CLEF.
Figure \ref{fig:parameters} shows the classification accuracies of tasks
I $\rightarrow$ P and P $\rightarrow$ I, by varying $\beta_1 \in \{1e-4, 1e-3, 1e-2, 1e-1,1e0\}$ and
$\beta_2 \in \{5e-6, 5e-5, 5e-4, 5e-3,5e-2,5e-1\}$.
We observe that the accuracy decreases slowly among the \textit{peak} area.
This confirms that MCI is stable enough under different parameter settings, which is vital
for the generality of an algorithm.

\begin{table}[t]
 \setlength{\abovecaptionskip}{0cm}
 \setlength{\belowcaptionskip}{-1cm}
    \caption{Dependence test on Office-31 task A$\rightarrow$D.
    Lower values indicate lower dependence.}
    \label{tab:independence_experiment}
    \vskip 0.08in
    \renewcommand{\tabcolsep}{0.28pc}
    \renewcommand{\arraystretch}{0.6}
    \begin{center}
    \begin{small}
    \begin{tabular}{c|ccc}
     \toprule
     Method &  $\hat{I}_n^{NOCCO}$ & $\hat{I}_{n_C}^{NOCCO}$& Accuracy(\%) \\
     \midrule
          Source \cite{he2016deep}       & 0.951 & 0.823 & 68.8  \\
          MCI (w/o $\mathcal{L}_{COND}$) & 0.517 & 0.664 & 89.9      \\
          DMP \cite{luo2020unsupervised} & 0.766 & 0.720 &  91.0 \\
          HSIC                           &\textbf{0.123} & 0.461 & 87.3    \\
          MCI                            & 0.280 & \textbf{0.398} & \textbf{92.2} \\
         \bottomrule
    \end{tabular}
    \end{small}
    \end{center}
    \vskip -0.15in
\end{table}

\textbf{Ablation Study}. To further explore the impact of $\mathcal{L}_{COND}$ and $\mathcal{L}_{Ent}$, we design ablation
experiments from three aspects:
modeling MCI without $\mathcal{L}_{COND}$,
modeling MCI without $\mathcal{L}_{Ent}$,
and modeling HSIC based on Eq.~\eqref{NOCCOcomparison}, which learns the
domain-invariant features by maximizing the independence
of the features and domain labels.
The ablation results are shown in Table \ref{tab:ablation_experiment}.
MCI consistently achieves the best, which suggests the class-conditioned transferring based on the conditional independence provides substantial advantages in UDA.
MCI (w/o $\mathcal{L}_{Ent}$) surpasses MCI (w/o $\mathcal{L}_{COND}$) with at least 1.6\% in accuracy, which indicates that loss $\mathcal{L}_{COND}$ plays a key role in the class-conditioned transferring.
The accuracies of MCI are higher than HSIC, which validates that the class-conditioned domain-invariant features are helpful
to train a discriminative classifier.
\begin{figure}[t]
\vskip 0.15in
\begin{center}
\includegraphics[width=175pt, trim= 20 250 20 280]{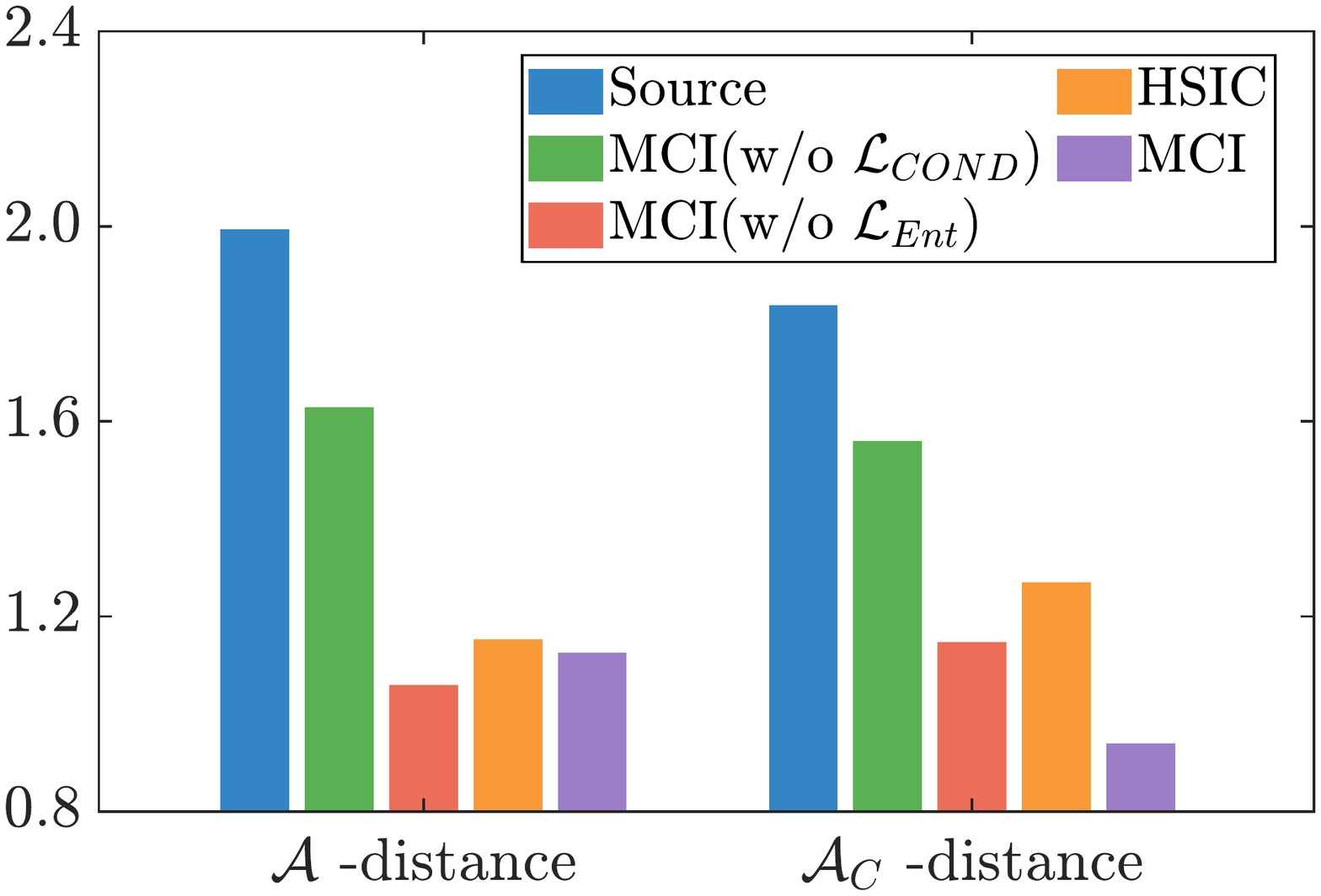}%
\caption{$\mathcal{A}$-distance and $\mathcal{A}_C$-distance
on Office-31 task A$\rightarrow$D. Best viewed in color.}
\label{fig:A_distance}
\end{center}
\vskip -0.2in
\end{figure}

\begin{table*}[t]
    \vskip 0.018in
    \caption{Accuracies (\%) on Office-10 (AlexNet), Image-CLEF, Office-31 and Office-Home (ResNet-50).}
    \label{tab:MS-MCI results on four datasets.}
    \vskip -0.1in
    \begin{center}
    \begin{small}

    \renewcommand{\tabcolsep}{0.3pc}
    \renewcommand{\arraystretch}{0.6}
    \begin{tabular}{m{2.0cm}<{\centering}|m{2.3cm}<{\centering}|m{1.2cm}<{\centering}m{1.2cm}<{\centering}m{1.2cm}<{\centering}m{1.2cm}<{\centering}|m{1.2cm}<{\centering}m{1.2cm}<{\centering}m{1.2cm}<{\centering}m{1.2cm}<{\centering}}
    \toprule
    \multirow{2}{*}{\textbf{Domain}} &  \multirow{2}{*}{\textbf{Method}} &\multicolumn{4}{c|}{\textbf{Image-CLEF}} &\multicolumn{4}{c}{\textbf{Office-31}}\\
    & &I,C$\rightarrow$P & I,P$\rightarrow$C & P,C$\rightarrow$I & Mean   &A,W$\rightarrow$D & A,D$\rightarrow$W & D,W$\rightarrow$A &  Mean  \\
    \midrule
    \multirow{5}{*}{Source-combine}& DAN \cite{long2018transferable} & 77.6 & 93.3 & 92.2 & 87.7 & 99.6 & 97.8 & 67.6 & 88.3 \\
                                   & D-CORAL \cite{sun2016deep}      & 77.1 & 93.6 & 91.7 & 87.5 & 99.3 & 98.0 & 67.1 & 88.1 \\
                                   & DANN \cite{ganin2016domain}  & 77.9 & 93.7 & 91.8 & 87.8 & 99.7 & 98.1 & 67.6 & 88.5 \\
                                   & DSAN \cite{zhu2020deep}  & 77.7 & 95.1 & 91.4 & 88.1 & 99.1 & 98.7 & 72.4 & 90.0\\
                                   & MCI                             & 80.8 & 95.1 & 91.4 & 89.1  & 99.7 & 96.3 & 70.4 & 88.8 \\
    \midrule
     \multirow{3}{*}{Multi-source}&  DCTN \cite{xu2018deep}       & 75.0 & 95.7 & 90.3 & 87.0       & 99.3 & 98.2 & 64.2 & 87.2 \\
                                   &  MFSAN \cite{zhu2019aligning} & 79.1 & 95.4 & \textbf{93.6} & 89.4      & 99.5 & 98.5 & 72.7 & 90.2  \\
                                   & MS-MCI       &\textbf{82.6} & \textbf{96.8} & \textbf{93.6} & \textbf{91.0} & \textbf{100.0}& \textbf{99.3} & \textbf{75.6} & \textbf{91.6} \\
    \bottomrule
    \end{tabular}
    \vskip 0.1in
    \renewcommand{\tabcolsep}{0.65pc}
    \renewcommand{\arraystretch}{0.6}
    \begin{tabular}{m{2.0cm}<{\centering}|m{2.5cm}<{\centering}|m{1.61cm}<{\centering}m{1.61cm}<{\centering}m{1.61cm}<{\centering}m{1.61cm}<{\centering}m{1.61cm}<{\centering}}
    \toprule
    \multirow{2}{*}{\textbf{Domain}} &  \multirow{2}{*}{\textbf{Method}} &\multicolumn{5}{c}{\textbf{Office-10}}\\
    & &A,C,D$\rightarrow$W & A,C,W$\rightarrow$D & A,D,W$\rightarrow$C & C,D,W$\rightarrow$A &  Mean \\
    \midrule
    \multirow{3}{*}{Source-combine}& DAN \cite{long2018transferable} & 99.3 & 98.2 & 89.7 & \textbf{94.8} & 95.5 \\
                                   & DSAN \cite{zhu2020deep}         & 98.3 & 99.4 & 86.4 & 92.5 & 94.2 \\
                                   & MCI                             & 98.5 & 99.4  & 86.6   & 93.4 & 94.5  \\
    \midrule
      \multirow{4}{*}{Multi-source}&  DCTN \cite{xu2018deep}        & 99.4 & 99.0 & 90.2 & 92.7 & 95.3 \\
                                   & MDAN \cite{zhao2018adversarial}  & 98.1& 98.2 & 89.5 & 92.2 & 94.5 \\
                                   & $\text{M}^3$SDA \cite{peng2019moment}& \textbf{99.5} & 99.2 & \textbf{91.5} & 94.1 & \textbf{96.1} \\
                                   & MS-MCI       & 99.2 &  \textbf{100.0} & 88.5 & 94.1 & 95.5 \\
    \bottomrule
    \end{tabular}
    \vskip 0.1in
    \renewcommand{\tabcolsep}{0.4pc}
    \renewcommand{\arraystretch}{0.6}
    \begin{tabular}{m{2.3cm}<{\centering}|m{2.3cm}<{\centering}|m{1.9cm}<{\centering}m{1.9cm}<{\centering}m{1.9cm}<{\centering}m{1.9cm}<{\centering}m{1.9cm}<{\centering}}
    \toprule
    \multirow{2}{*}{\textbf{Domain}} &  \multirow{2}{*}{\textbf{Method}} &\multicolumn{5}{c}{\textbf{Office-Home}}\\
    & &Cl,Pr,Rw$\rightarrow$Art & Ar,Pr,Rw$\rightarrow$Cl & Ar,Cl,Rw$\rightarrow$Pr & Ar,Cl,Pr,$\rightarrow$Rw &  Mean \\
    \midrule
    \multirow{5}{*}{Source-combine}& DAN \cite{long2018transferable}     & 68.5 & 59.4 & 79.0 & 82.5 & 72.4 \\
                                   & D-CORAL \cite{sun2016deep}          & 68.1 & 58.6 & 79.5 & 82.7 & 72.2\\
                                   & DANN \cite{ganin2016domain}         & 68.4 & 59.1 & 79.5 & 82.7 & 72.4 \\
                                   & DSAN \cite{zhu2020deep}             & 69.1 & 57.7 & 72.5 & 73.0 & 68.1 \\
                                   & MCI                            & 70.1  &56.1& \textbf{84.8} & 82.5 & 73.4   \\
    \midrule
    \multirow{5}{*}{Multi-source} & MDAN \cite{zhao2018adversarial}  & 68.1& 67.0 & 81.0 &82.8 & 74.8 \\
                                   &  MDMN \cite{li2018extracting} & 68.7&\textbf{67.8}& 81.4 &83.3 & \textbf{75.3} \\
                                   &  $\text{M}^3$SDA \cite{peng2019moment} & 64.1 & 62.8& 76.2 & 78.6 & 70.4 \\
                                   & MFSAN \cite{zhu2019aligning} & \textbf{72.1} & 62.0 & 80.3 & 81.8 & 74.1 \\
                                   & MS-MCI       & 70.1 & 58.1  & 84.7 & \textbf{83.9}  & 74.2 \\
    \bottomrule
    \end{tabular}
    \end{small}
    \end{center}
    \vskip -0.2in
  \end{table*}

\textbf{Dependence Test}. To explore the dependence of feature and domain,
we perform the empirical estimation $\hat{I}_n^{NOCCO}$ on Office-31 task A$\rightarrow$D.
We also define $\hat{I}_{n_C}^{NOCCO} = \mathbb{E}[\hat{I}_{n_c}^{NOCCO}]$ to estimate the dependence of the features and domain labels at class-level, where $\hat{I}_{n_c}^{NOCCO}$ is the $\hat{I}_n^{NOCCO}$ of features based on class $\mathbf{y}_c$.
Results are shown in Table~\ref{tab:independence_experiment}.
Since DMP aims to explore the discriminative structure of the target domain based on the manifold embedding,
it is natural to observed that DMP has higher $\hat{I}_n^{NOCCO}$ and $\hat{I}_{n_c}^{NOCCO}$ than MCI.
MCI even achieves a better result than DMP on this transfer task, which further validates that MCI provides a novel insight to deal with
domain adaptation. As MCI (w/o $\mathcal{L}_{COND}$) improves the classifier by exploring the entropy criterion, it has a strong dependence between feature and domain.
Though the independence based method HSIC has the lowest $\hat{I}_n^{NOCCO}$, its classification accuracy is worse than MCI (w/o $\mathcal{L}_{COND}$).
MCI achieves the highest accuracy along with the lowest $\hat{I}_{n_C}^{NOCCO}$. This indicates that considering class-conditioned information plays a vital role in UDA.
Interestingly, the $\hat{I}_{n_C}^{NOCCO}$ of MCI is larger than its $\hat{I}_n^{NOCCO}$, which confirms that learning class-conditioned domain-invariant features is more difficult than the domain-invariant ones.

\textbf{Feature Visualization}. We visualize the 2-D representations before and after adaptation by t-SNE \cite{maaten2008visualizing}.
The visualization results of Office-31 task A$\rightarrow$D are shown in Figure~\ref{fig:tsne}.
It is observed that the source and target domains have different spatial distributions before adaptation.
Figure~\ref{fig:office-31-before-class} validates that it is hard to classify the target samples with the classifier trained only
on the labeled source domain. As expected, the source and target domains have obvious
cluster structures after adaptation by MCI.
Figure~\ref{fig:office-31-after-class} further validates that MCI achieves the class-conditional distribution alignment well.

\textbf{Distribution Discrepancy}.
The $\mathcal{A}$-distance \cite{ben2010theory} is usually used to measure the distribution discrepancy between domains.
The global distribution discrepancy is estimated by $d_{\mathcal{A}} = 2(1-2\epsilon)$, where $\epsilon$ is the test error
of a classifier which is trained to discriminate the source and target domains.
We also estimate the class-conditional distribution discrepancy by $d_{\mathcal{A}_C} = \mathbb{E}[d_{\mathcal{A}_c}]$,
where $d_{\mathcal{A}_c}$ is the $\mathcal{A}$-distance of the class-conditional distributions based on class $\mathbf{y}_c$. More details are
described in \cite{zhu2020deep}.
Results are shown in Figure~\ref{fig:A_distance}.
Both the $\mathcal{A}$-distance and $\mathcal{A}_C$-distance of MCI (w/o $\mathcal{L}_{COND}$) are smaller than MCI (w/o $\mathcal{L}_{Ent}$), which further validates that
loss $\mathcal{L}_{COND}$ is the key of MCI.
Though HSIC and MCI have similar $\mathcal{A}$-distance, the
$\mathcal{A}_C$-distance of MCI is much smaller than HSIC.
Obviously, MCI is helpful to learn more separable features by achieving the class-conditioned transferring.

\subsection{Results and Analysis for MS-MCI}
\textbf{Results}. The classification accuracy on four datasets are shown in Table~\ref{tab:MS-MCI results on four datasets.}. ``Source-combine" means combine all the source domains into a single-source domain,
and then the multi-source scenario changes to a traditional single-source scenario. ``Multi-source" denotes all the source domains are employed to train a classifier for the target domain.

The classification accuracies on Image-CLEF are shown in the left top of Table~\ref{tab:MS-MCI results on four datasets.}.
MS-MCI outperforms other source-combine and multi-source methods on all the transfer tasks. MCI improves 1.6\% than the second best model MFSAN \cite{zhu2019aligning}, which validates that consider the
domain-specific information is necessary in domain adaptation.

The classification accuracies on Office-31 are shown in the right top of Table~\ref{tab:MS-MCI results on four datasets.}.
MS-MCI again exceeds other methods on all transfer tasks. The accuracy of MCI achieves \textbf{100\%} on task A,W$\rightarrow$D, and increases by 2.9\% on task D,W$\rightarrow$A, which further validate the effectiveness of
extending MCI to MDA.

The classification accuracy on Office-10 are shown in the second row of Table~\ref{tab:MS-MCI results on four datasets.}. It is observed that MS-MCI achieves \textbf{100\%} on task A,C,W$\rightarrow$D.

The classification accuracy of Office-Home are shown in the bottom of Table~\ref{tab:MS-MCI results on four datasets.}. Since there are more classes and larger domain discrepancy in the multi-source scenario
of Office-Home, it is more challenging to learn the class-conditioned domain-invariant representations than other datasets. Compared with MFSAN \cite{zhu2019aligning}, the accuracy of MS-MCI increases by 2.1\% on task
Ar,Pr,Rw$\rightarrow$Pr and the average accuracy is 74.2\% , which is a slightly lower but has reached the most advanced methods.

\textbf{Feature Visualization}. We visualize the 2-D representations of MFSAN and MS-MCI by t-SNE. The visualization results of Office-31 task A,D$\rightarrow$W are shown in Figure~\ref{fig:tsne-MS-MCI}. We can observe that MFSAN achieves the intra-class compactness on domain D and W while loses the inter-class separability on domain A. It is obvious that MS-MCI can benefit from the conditional independence: scatters from the same class but different domains are nearby in Figure~\ref{fig:MS-MCI-office-31-after-class}, which further validates that MS-MCI achieves the class-conditional distribution alignment well.

\section{Conclusion}
\label{conclusion}
In this paper, we deal with UDA by removing the domain-specific information while preserving discriminative structure simultaneously. Specifically, we explore the class-conditioned transferring from a statistical perspective, which is maximizing the conditional independence of the extracted features and domain-specific information. Meanwhile, this transferring derives a class-conditional distribution alignment mathematically. By providing an interpretable empirical estimation of the conditional dependence, it is clear that the class-conditional information is sufficiently considered to learn the class-conditioned
domain-invariant features. MCI can be adopted in both the single-source and multi-source scenarios. We also derive informative upper bounds of the target error based on the class-conditional distributions, which provide theoretical insights of our proposal under both scenarios. Extensive experiments demonstrate the effectiveness of solving the domain adaptation problem from a statistical conditional independence.
\ifCLASSOPTIONcaptionsoff
  \newpage
\fi

\bibliography{MCI}

\begin{thebibliography}{10}
\providecommand{\url}[1]{#1}
\csname url@samestyle\endcsname
\providecommand{\newblock}{\relax}
\providecommand{\bibinfo}[2]{#2}
\providecommand{\BIBentrySTDinterwordspacing}{\spaceskip=0pt\relax}
\providecommand{\BIBentryALTinterwordstretchfactor}{4}
\providecommand{\BIBentryALTinterwordspacing}{\spaceskip=\fontdimen2\font plus
\BIBentryALTinterwordstretchfactor\fontdimen3\font minus
  \fontdimen4\font\relax}
\providecommand{\BIBforeignlanguage}[2]{{%
\expandafter\ifx\csname l@#1\endcsname\relax
\typeout{** WARNING: IEEEtran.bst: No hyphenation pattern has been}%
\typeout{** loaded for the language `#1'. Using the pattern for}%
\typeout{** the default language instead.}%
\else
\language=\csname l@#1\endcsname
\fi
#2}}
\providecommand{\BIBdecl}{\relax}
\BIBdecl

\bibitem{long2018transferable}
M.~Long, Y.~Cao, Z.~Cao, J.~Wang, and M.~I. Jordan, ``Transferable
  representation learning with deep adaptation networks,'' \emph{IEEE
  Transactions on Pattern Analysis and Machine Intelligence}, vol.~41, no.~12,
  pp. 3071--3085, 2019.

\bibitem{pan2009survey}
S.~J. Pan and Q.~Yang, ``A survey on transfer learning,'' \emph{IEEE
  Transactions on Knowledge and Data Engineering}, vol.~22, no.~10, pp.
  1345--1359, 2010.

\bibitem{pan2010domain}
S.~J. Pan, I.~W. Tsang, J.~T. Kwok, and Q.~Yang, ``Domain adaptation via
  transfer component analysis,'' \emph{IEEE Transactions on Neural Networks},
  vol.~22, no.~2, pp. 199--210, 2010.

\bibitem{Ren_TSCDA}
C.-X. Ren, P.~Ge, P.~Yang, and S.~Yan, ``Learning target-domain-specific
  classifier for partial domain adaptation,'' \emph{IEEE Transactions on Neural
  Networks and Learning Systems}, vol.~32, no.~5, pp. 1989--2001, 2021.

\bibitem{lin2016cross}
Y.~Lin, J.~Chen, Y.~Cao, Y.~Zhou, L.~Zhang, Y.~Y. Tang, and S.~Wang,
  ``Cross-domain recognition by identifying joint subspaces of source domain
  and target domain,'' \emph{IEEE Transactions on Cybernetics}, vol.~47, no.~4,
  pp. 1090--1101, 2017.

\bibitem{Khodabandeh_2019_ICCV}
M.~Khodabandeh, A.~Vahdat, M.~Ranjbar, and W.~G. Macready, ``A robust learning
  approach to domain adaptive object detection,'' in \emph{Proceedings of the
  IEEE International Conference on Computer Vision}, October 2019.

\bibitem{Khur2021}
S.~Khurana, N.~Moritz, T.~Hori, and J.~L. Roux, ``Unsupervised domain
  adaptation for speech recognition via uncertainty driven self-training,'' in
  \emph{ICASSP 2021 - 2021 IEEE International Conference on Acoustics, Speech
  and Signal Processing}, 2021, pp. 6553--6557.

\bibitem{Xu_FLARE}
G.-X. Xu, C.~Liu, J.~Liu, Z.~Ding, F.~Shi, M.~Guo, W.~Zhao, X.~Li, Y.~Wei,
  Y.~Gao, C.-X. Ren, and D.~Shen, ``Cross-site severity assessment of covid-19
  from ct images via domain adaptation,'' \emph{IEEE Transactions on Medical
  Imaging}, vol.~41, no.~1, pp. 88--102, 2022.

\bibitem{ben2007analysis}
S.~Ben-David, J.~Blitzer, K.~Crammer, and F.~Pereira, ``Analysis of
  representations for domain adaptation,'' in \emph{Advances in Neural
  Information Processing Systems}, vol.~19, 2007.

\bibitem{long2013transfer}
M.~Long, J.~Wang, G.~Ding, J.~Sun, and P.~S. Yu, ``Transfer feature learning
  with joint distribution adaptation,'' in \emph{Proceedings of the IEEE
  International Conference on Computer Vision}, December 2013.

\bibitem{gong2012geodesic}
B.~Gong, Y.~Shi, F.~Sha, and K.~Grauman, ``Geodesic flow kernel for
  unsupervised domain adaptation,'' in \emph{Proceedings of the IEEE Conference
  on Computer Vision and Pattern Recognition}, 2012, pp. 2066--2073.

\bibitem{ren2019heterogeneous}
C.~X. Ren, J.~Feng, D.~Q. Dai, and S.~Yan, ``Heterogeneous domain adaptation
  via covariance structured feature translators,'' \emph{IEEE Transactions on
  Cybernetics}, vol.~51, no.~4, pp. 2166--2177, 2021.

\bibitem{luo2020unsupervised}
Y.~W. Luo, C.~X. Ren, D.~Q. DAI, and H.~Yan, ``Unsupervised domain adaptation
  via discriminative manifold propagation,'' \emph{IEEE Transactions on Pattern
  Analysis and Machine Intelligence}, pp. 1--1, 2020.

\bibitem{yosinski2014transferable}
J.~Yosinski, J.~Clune, Y.~Bengio, and H.~Lipson, ``How transferable are
  features in deep neural networks?'' in \emph{Advances in Neural Information
  Processing Systems}, vol.~27, 2014.

\bibitem{ganin2016domain}
Y.~Ganin, E.~Ustinova, H.~Ajakan, P.~Germain, H.~Larochelle, F.~Laviolette,
  M.~Marchand, and V.~Lempitsky, ``Domain-adversarial training of neural
  networks,'' \emph{The Journal of Machine Learning Research}, vol.~17, no.~1,
  pp. 2096--2030, 2016.

\bibitem{tzeng2017adversarial}
E.~Tzeng, J.~Hoffman, K.~Saenko, and T.~Darrell, ``Adversarial discriminative
  domain adaptation,'' in \emph{Proceedings of the IEEE Conference on Computer
  Vision and Pattern Recognition}, July 2017.

\bibitem{ren2019domain}
C.~X. Ren, B.~Liang, P.~Ge, Y.~Zhai, and Z.~Lei, ``Domain adaptive person
  re-identification via camera style generation and label propagation,''
  \emph{IEEE Transactions on Information Forensics and Security}, vol.~15, pp.
  1290--1302, 2020.

\bibitem{pan2019transferrable}
Y.~Pan, T.~Yao, Y.~Li, Y.~Wang, C.-W. Ngo, and T.~Mei, ``Transferrable
  prototypical networks for unsupervised domain adaptation,'' in
  \emph{Proceedings of the IEEE Conference on Computer Vision and Pattern
  Recognition}, 2019, pp. 2239--2247.

\bibitem{liang2019distant}
J.~Liang, R.~He, Z.~Sun, and T.~Tan, ``Distant supervised centroid shift: A
  simple and efficient approach to visual domain adaptation,'' in
  \emph{Proceedings of the IEEE Conference on Computer Vision and Pattern
  Recognition}, 2019, pp. 2975--2984.

\bibitem{Luo_2021_CVPR}
Y.~W. Luo and C.~X. Ren, ``Conditional bures metric for domain adaptation,'' in
  \emph{Proceedings of the IEEE Conference on Computer Vision and Pattern
  Recognition}, June 2021, pp. 13\,989--13\,998.

\bibitem{sun2015survey}
S.~Sun, H.~Shi, and Y.~Wu, ``A survey of multi-source domain adaptation,''
  \emph{Information Fusion}, vol.~24, pp. 84--92, 2015.

\bibitem{riemer2018learning}
M.~Riemer, I.~Cases, R.~Ajemian, M.~Liu, I.~Rish, Y.~Tu, , and G.~Tesauro,
  ``Learning to learn without forgetting by maximizing transfer and minimizing
  interference,'' in \emph{International Conference on Learning
  Representations}, 2019.

\bibitem{fukumizu2007kernel}
K.~Fukumizu, A.~Gretton, X.~Sun, and B.~Sch\"{o}lkopf, ``Kernel measures of
  conditional dependence,'' in \emph{Advances in Neural Information Processing
  Systems}, vol.~20, 2008.

\bibitem{sun2016return}
B.~Sun, J.~Feng, and K.~Saenko, ``Return of frustratingly easy domain
  adaptation,'' in \emph{Proceedings of the AAAI Conference on Artificial
  Intelligence}, vol.~30, no.~1, 2016.

\bibitem{Yan_2017_CVPR}
H.~Yan, Y.~Ding, P.~Li, Q.~Wang, Y.~Xu, and W.~Zuo, ``Mind the class weight
  bias: Weighted maximum mean discrepancy for unsupervised domain adaptation,''
  in \emph{Proceedings of the IEEE Conference on Computer Vision and Pattern
  Recognition}, July 2017.

\bibitem{courty2016optimal}
N.~Courty, R.~Flamary, D.~Tuia, and A.~Rakotomamonjy, ``Optimal transport for
  domain adaptation,'' \emph{IEEE Transactions on Pattern Analysis and Machine
  Intelligence}, vol.~39, no.~9, pp. 1853--1865, 2017.

\bibitem{li2020enhanced}
M.~Li, Y.~M. Zhai, Y.~W. Luo, P.~F. Ge, and C.~X. Ren, ``Enhanced transport
  distance for unsupervised domain adaptation,'' in \emph{Proceedings of the
  IEEE Conference on Computer Vision and Pattern Recognition}, June 2020.

\bibitem{zhang2019optimal}
Z.~Zhang, M.~Wang, and A.~Nehorai, ``Optimal transport in reproducing kernel
  hilbert spaces: Theory and applications,'' \emph{IEEE Transactions on Pattern
  Analysis and Machine Intelligence}, vol.~42, no.~7, pp. 1741--1754, 2020.

\bibitem{long2018conditional}
M.~Long, Z.~Cao, J.~Wang, and M.~I. Jordan, ``Conditional adversarial domain
  adaptation,'' in \emph{Advances in neural information processing systems},
  2018, pp. 1640--1650.

\bibitem{jiang2020implicit}
X.~Jiang, Q.~Lao, S.~Matwin, and M.~Havaei, ``Implicit class-conditioned domain
  alignment for unsupervised domain adaptation,'' in \emph{Proceedings of the
  37th International Conference on Machine Learning}, vol. 119, 13--18 Jul
  2020, pp. 4816--4827.

\bibitem{xie2018learning}
S.~Xie, Z.~Zheng, L.~Chen, and C.~Chen, ``Learning semantic representations for
  unsupervised domain adaptation,'' in \emph{International Conference on
  Machine Learning}, 2018, pp. 5423--5432.

\bibitem{deng2019cluster}
Z.~Deng, Y.~Luo, and J.~Zhu, ``Cluster alignment with a teacher for
  unsupervised domain adaptation,'' in \emph{Proceedings of the IEEE
  International Conference on Computer Vision}, 2019, pp. 9944--9953.

\bibitem{zhao2019learning}
H.~Zhao, R.~T.~D. Combes, K.~Zhang, and G.~Gordon, ``On learning invariant
  representations for domain adaptation,'' in \emph{Proceedings of the 36th
  International Conference on Machine Learning}, vol.~97, 09--15 Jun 2019, pp.
  7523--7532.

\bibitem{zhu2020deep}
Y.~Zhu, F.~Zhuang, J.~Wang, G.~Ke, J.~Chen, J.~Bian, H.~Xiong, and Q.~He,
  ``Deep subdomain adaptation network for image classification,'' \emph{IEEE
  Transactions on Neural Networks and Learning Systems}, 2020.

\bibitem{Ren2021}
C.-X. Ren, P.~Ge, D.-Q. Dai, and H.~Yan, ``Learning kernel for conditional
  moment-matching discrepancy-based image classification,'' \emph{IEEE
  Transactions on Cybernetics}, vol.~51, no.~4, pp. 2006--2018, 2021.

\bibitem{Ke2018}
K.~Yan, L.~Kou, and D.~Zhang, ``Learning domain-invariant subspace using domain
  features and independence maximization,'' \emph{IEEE Transactions on
  Cybernetics}, vol.~48, no.~1, pp. 288--299, 2018.

\bibitem{blitzer2007learning}
J.~Blitzer, K.~Crammer, A.~Kulesza, F.~Pereira, and J.~Wortman, ``Learning
  bounds for domain adaptation,'' in \emph{Advances in Neural Information
  Processing Systems}, vol.~20, 2008.

\bibitem{ben2010theory}
S.~Ben-David, J.~Blitzer, K.~Crammer, A.~Kulesza, F.~Pereira, and J.~W.
  Vaughan, ``A theory of learning from different domains,'' \emph{Machine
  learning}, vol.~79, no. 1-2, pp. 151--175, 2010.

\bibitem{mansour2009domain}
Y.~Mansour, M.~Mohri, and A.~Rostamizadeh, ``Domain adaptation with multiple
  sources,'' in \emph{Advances in Neural Information Processing Systems},
  vol.~21, 2009.

\bibitem{xu2018deep}
R.~Xu, Z.~Chen, W.~Zuo, J.~Yan, and L.~Lin, ``Deep cocktail network:
  Multi-source unsupervised domain adaptation with category shift,'' in
  \emph{Proceedings of the IEEE Conference on Computer Vision and Pattern
  Recognition}, June 2018.

\bibitem{Liu_TWMDA}
Y.-H. Liu and C.-X. Ren, ``A two-way alignment approach for unsupervised
  multi-source domain adaptation,'' \emph{Pattern Recognition}, vol. 124, p.
  108430, 2022.

\bibitem{PTMDA_Ren2022}
C.-X. Ren, Y.-H. Liu, X.-W. Zhang, and K.-K. Huang, ``Multi-source unsupervised
  domain adaptation via pseudo target domain,'' \emph{IEEE Transactions on
  Image Processing}, vol.~31, pp. 2122--2135, 2022.

\bibitem{li2018extracting}
Y.~Li, M.~Murias, S.~Major, G.~Dawson, and D.~E. Carlson, ``Extracting
  relationships by multi-domain matching,'' in \emph{Proceedings of the 32nd
  International Conference on Neural Information Processing Systems}, 2018, pp.
  6799--6810.

\bibitem{peng2019moment}
X.~Peng, Q.~Bai, X.~Xia, Z.~Huang, K.~Saenko, and B.~Wang, ``Moment matching
  for multi-source domain adaptation,'' in \emph{Proceedings of the IEEE
  International Conference on Computer Vision}, October 2019.

\bibitem{zhao2018adversarial}
H.~Zhao, S.~Zhang, G.~Wu, J.~M.~F. Moura, J.~P. Costeira, and G.~J. Gordon,
  ``Adversarial multiple source domain adaptation,'' in \emph{Advances in
  Neural Information Processing Systems}, vol.~31, 2018.

\bibitem{wen2020domain}
J.~Wen, R.~Greiner, and D.~Schuurmans, ``Domain aggregation networks for
  multi-source domain adaptation,'' in \emph{Proceedings of the 37th
  International Conference on Machine Learning}, vol. 119, 13--18 Jul 2020, pp.
  10\,214--10\,224.

\bibitem{zhu2019aligning}
Y.~Zhu, F.~Zhuang, and D.~Wang, ``Aligning domain-specific distribution and
  classifier for cross-domain classification from multiple sources,'' in
  \emph{Proceedings of the AAAI Conference on Artificial Intelligence},
  vol.~33, no.~01, 2019, pp. 5989--5996.

\bibitem{baker1973joint}
C.~R. Baker, ``Joint measures and cross-covariance operators,''
  \emph{Transactions of the American Mathematical Society}, vol. 186, pp.
  273--289, 1973.

\bibitem{fukumizu2007statistical}
K.~Fukumizu, F.~R. Bach, and A.~Gretton, ``Statistical consistency of kernel
  canonical correlation analysis,'' \emph{Journal of Machine Learning
  Research}, vol.~8, no. Feb, pp. 361--383, 2007.

\bibitem{fukumizu2004dimensionality}
K.~Fukumizu, F.~R. Bach, and M.~I. Jordan, ``Dimensionality reduction for
  supervised learning with reproducing kernel hilbert spaces,'' \emph{Journal
  of Machine Learning Research}, vol.~5, no. Jan, pp. 73--99, 2004.

\bibitem{fukumizu2009kernel}
K.~Fukumizu, F.~R. Bach, M.~I. Jordan \emph{et~al.}, ``Kernel dimension
  reduction in regression,'' \emph{The Annals of Statistics}, vol.~37, no.~4,
  pp. 1871--1905, 2009.

\bibitem{gretton2005measuring}
A.~Gretton, O.~Bousquet, A.~Smola, and B.~Sch{\"o}lkopf, ``Measuring
  statistical dependence with hilbert-schmidt norms,'' in \emph{International
  Conference on Algorithmic Learning Theory}, 2005, pp. 63--77.

\bibitem{crammer2008learning}
K.~Crammer, M.~Kearns, and J.~Wortman, ``Learning from multiple sources,''
  \emph{Journal of Machine Learning Research}, vol.~9, no.~57, pp. 1757--1774,
  2008.

\bibitem{he2016deep}
K.~He, X.~Zhang, S.~Ren, and J.~Sun, ``Deep residual learning for image
  recognition,'' in \emph{Proceedings of the IEEE Conference on Computer Vision
  and Pattern Recognition}, June 2016.

\bibitem{xu2019larger}
R.~Xu, G.~Li, J.~Yang, and L.~Lin, ``Larger norm more transferable: An adaptive
  feature norm approach for unsupervised domain adaptation,'' in
  \emph{Proceedings of the IEEE International Conference on Computer Vision},
  October 2019.

\bibitem{krizhevsky2017imagenet}
A.~Krizhevsky, I.~Sutskever, and G.~E. Hinton, ``Imagenet classification with
  deep convolutional neural networks,'' \emph{Communications of the ACM},
  vol.~60, no.~6, pp. 84--90, 2017.

\bibitem{caputo2014imageclef}
B.~Caputo, H.~M{\"u}ller, J.~Martinez-Gomez, M.~Villegas, B.~Acar, N.~Patricia,
  N.~Marvasti, S.~{\"U}sk{\"u}darl{\i}, R.~Paredes, M.~Cazorla \emph{et~al.},
  ``Imageclef 2014: Overview and analysis of the results,'' in
  \emph{International Conference of the Cross-Language Evaluation Forum for
  European Languages}, 2014, pp. 192--211.

\bibitem{saenko2010adapting}
K.~Saenko, B.~Kulis, M.~Fritz, and T.~Darrell, ``Adapting visual category
  models to new domains,'' in \emph{European Conference on Computer Vision},
  2010, pp. 213--226.

\bibitem{griffin2007caltech}
G.~Griffin, A.~Holub, and P.~Perona, ``Caltech-256 object category dataset,''
  2007.

\bibitem{venkateswara2017deep}
H.~Venkateswara, J.~Eusebio, S.~Chakraborty, and S.~Panchanathan, ``Deep
  hashing network for unsupervised domain adaptation,'' in \emph{Proceedings of
  the IEEE Conference on Computer Vision and Pattern Recognition}, July 2017.

\bibitem{wang2020unsupervised}
Q.~Wang and T.~Breckon, ``Unsupervised domain adaptation via structured
  prediction based selective pseudo-labeling,'' in \emph{Proceedings of the
  AAAI Conference on Artificial Intelligence}, vol.~34, no.~04, 2020, pp.
  6243--6250.

\bibitem{paszke2019pytorch}
A.~Paszke, S.~Gross, F.~Massa, A.~Lerer, J.~Bradbury, G.~Chanan, T.~Killeen,
  Z.~Lin, N.~Gimelshein, L.~Antiga, A.~Desmaison, A.~Kopf, E.~Yang, Z.~DeVito,
  M.~Raison, A.~Tejani, S.~Chilamkurthy, B.~Steiner, L.~Fang, J.~Bai, and
  S.~Chintala, ``Pytorch: An imperative style, high-performance deep learning
  library,'' in \emph{Advances in Neural Information Processing Systems},
  vol.~32, 2019.

\bibitem{sun2016deep}
B.~Sun and K.~Saenko, ``Deep coral: Correlation alignment for deep domain
  adaptation,'' in \emph{European Conference on Computer Vision}.\hskip 1em
  plus 0.5em minus 0.4em\relax Springer, 2016, pp. 443--450.

\bibitem{maaten2008visualizing}
L.~v.~d. Maaten and G.~Hinton, ``Visualizing data using t-sne,'' \emph{Journal
  of Machine Learning Research}, vol.~9, no. Nov, pp. 2579--2605, 2008.

\end{thebibliography}
\bibliographystyle{IEEEtran}
\end{document}